\documentclass{article}

\usepackage[nonatbib,preprint]{neurips_2026}

\usepackage[numbers,sort]{natbib}
\usepackage[utf8]{inputenc} % allow utf-8 input
\usepackage[T1]{fontenc}    % use 8-bit T1 fonts
\usepackage{hyperref}       % hyperlinks
\usepackage{url}            % simple URL typesetting
\usepackage{booktabs}       % professional-quality tables
\usepackage{amsfonts}       % blackboard math symbols
\usepackage{nicefrac}       % compact symbols for 1/2, etc.
\usepackage{xcolor,colortbl}         % colors
\usepackage[inline]{enumitem}
\usepackage{graphicx}    % For \resizebox
\usepackage{makecell}    % For \makecell
\usepackage{multirow}
\usepackage{overpic}
\usepackage{float}
\usepackage{wrapfig}
\usepackage{amsmath}
\usepackage{subcaption}
\usepackage{pgfplots}
\usepackage{amsthm}
\usepackage{tcolorbox}
\usepgfplotslibrary{statistics}
\pgfplotsset{compat=1.17}
\tcbuselibrary{breakable, listings}

\makeatletter
\renewcommand{\paragraph}{%
  \@startsection{paragraph}{4}{\z@}%
                {0.0ex \@plus 0.3ex \@minus 0.1ex}%
                {-1em}%
                {\normalsize\bf}%
}
\makeatother

\newlength\savewidth\newcommand\shline{\noalign{\global\savewidth\arrayrulewidth
\global\arrayrulewidth 1pt}\hline\noalign{\global\arrayrulewidth\savewidth}}

\newcommand{\tablestyle}[2]{\setlength{\tabcolsep}{#1}\renewcommand{\arraystretch}{#2}\centering\footnotesize}

\def \dataset {{OrionData}}
\def \etal {{\emph{et al}.}}
\def \eg {{\emph{e.g}.\thinspace}}

\def \alg {{EleRPO}}
\def \model {{DAD}}
\def \dataset {{DAD-10M}}

\title{Detect Anything in Graphic Design: \\Element-Level Rewards for Autoregressive Detection}

\author{Jiangning Zhu$^{1}$, Bowen Li$^{1}$, Shenyu Qiao$^{1}$, Yima Gu$^{1}$ \\\textbf{Zhao Zhang$^{2}$, Yuhui Yuan$^{2}$\hspace{1.5mm}, Shixia Liu}$^{1}$\thanks{Corresponding author.}\\
	$^1$BNRist, Tsinghua University $^2$Canva Research\\
}

\begin{document}

\maketitle

% \begin{figure}[H]
% \centering
% \includegraphics[width=1\linewidth]{figs/teaser.pdf}
% % \caption{An overview of three applications that demonstrate the usefulness of \dataset{}.}
% \caption{\footnotesize Key Contributions: (i) An open-source benchmark \dataset{}. (ii) Improvements on chart understanding, infographics object detection, and graphic layout detection.}
% \label{fig:teaser}
% \vspace{-4mm}
% \end{figure}

\begin{figure}[H]
\centering
    \centering
    \includegraphics[width=\linewidth]{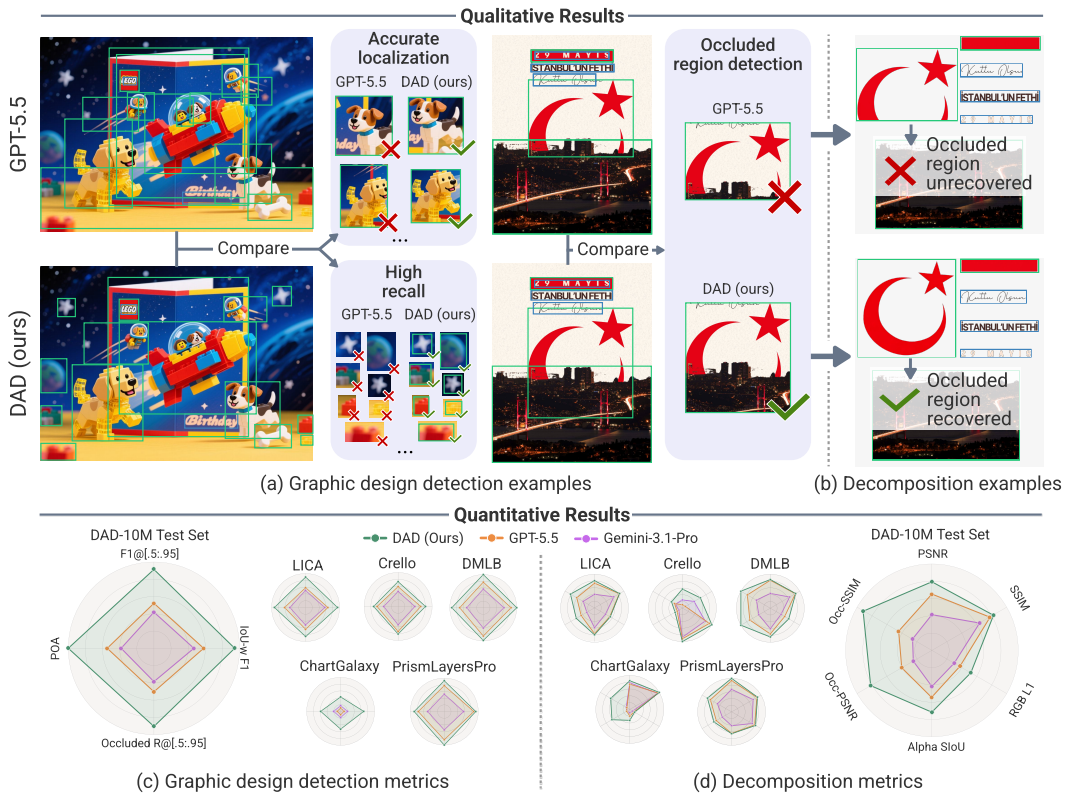}
    \caption{The \model{} model detects elements missed by GPT-5.5 and recovers occluded regions.
    It performs the best across all datasets and metrics.
    Metrics are normalized to $[0.1,1]$ in radar charts.\looseness=-1
    }
\label{fig:qualitative}
\vspace{-4mm}
\end{figure}

\begin{abstract}
% v1

Graphic designs, such as posters, advertisements, and infographics, are an important medium for communicating information and shaping understanding.
Unlike natural images, they consist of layered elements with explicit compositional order.
However, existing object detection models treat these elements as an unordered set, leaving compositional order unexploited.
To address this limitation, we present \textbf{D}etect \textbf{A}nything in Graphic \textbf{D}esign (DAD), a model that formulates graphic design detection as compositional deconstruction. 
It decodes elements in compositional order, using lower-layer elements to better detect higher-layer ones.
The key feature of DAD is amodal detection, which predicts the full bounding box of each element, including regions occluded by elements placed above it.
Building on this formulation, we propose \textbf{Ele}ment \textbf{R}elative \textbf{P}olicy \textbf{O}ptimization (EleRPO), which extends GRPO from sequence-level supervision to element-level optimization. 
EleRPO provides fine-grained training signals that capture how each detected element contributes to overall detection quality, and works synergistically with compositional order to improve detection performance.
To support training and evaluation, we build a dataset of 10 million graphic designs.
Experiments show that DAD outperforms all baselines and achieves human-level performance in amodal detection, supporting effective image-to-layer decomposition. 
EleRPO consistently improves over GRPO across nine detection benchmarks.

\raisebox{-0.3\height}{\hspace{0.05cm}\includegraphics[width=0.45cm]{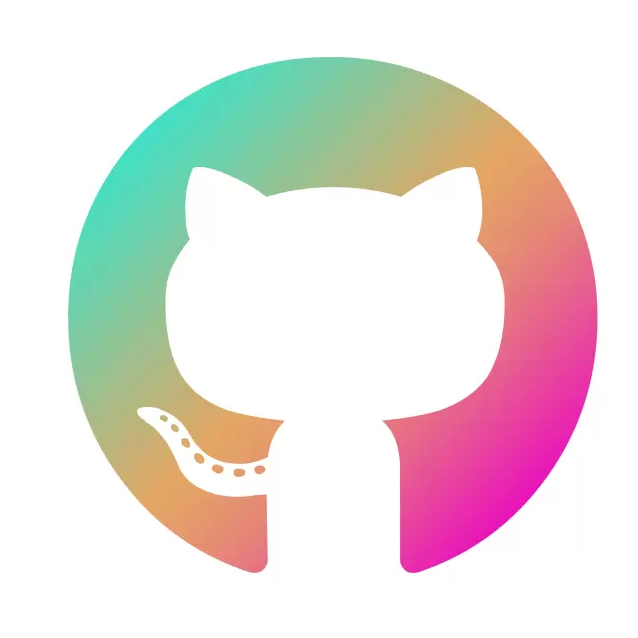}} \small \textbf{\mbox{Code:}} \href{https://github.com/dad887/DAD}{https://github.com/dad887/DAD} \\
\vspace{1em}
\raisebox{-0.3\height}{\includegraphics[width=0.4cm]{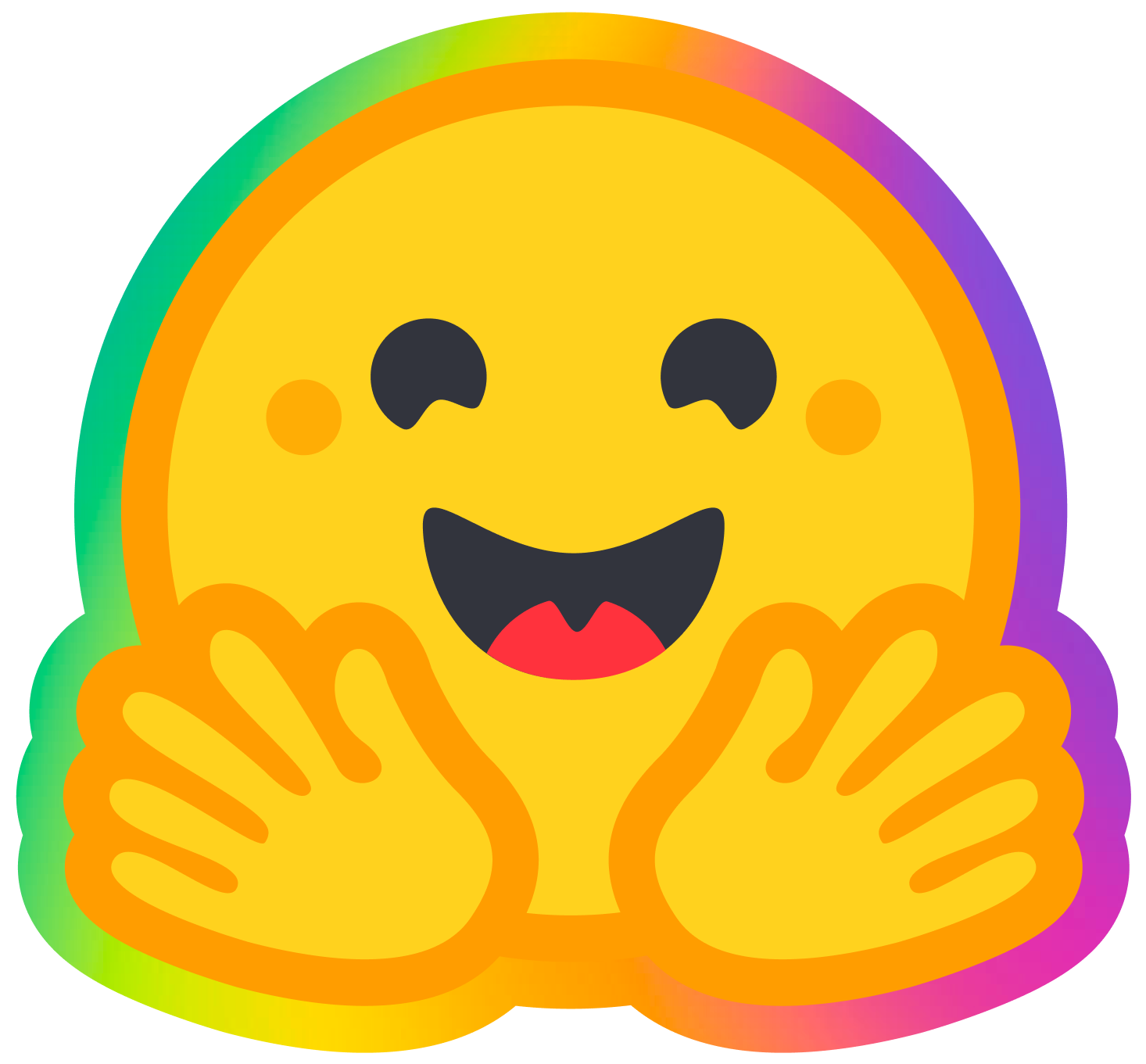}} \small \textbf{\mbox{Data \& Dataset Card:}} \href{https://huggingface.co/datasets/dad887/DAD}{https://huggingface.co/datasets/dad887/DAD}

\end{abstract}

\section{Introduction}
\label{sec:intro}

Graphic designs, such as posters, advertisements, and infographics, are a ubiquitous visual medium that shapes how people consume information across business communication, news media, public services, and education. 
By combining text with visual elements such as photographs, pictograms, and diagrams, they clearly organize complex information, attract attention, and make ideas easier to understand. 
A key property that distinguishes graphic designs from natural images is their compositional nature.
They are typically built from discrete, layered elements that are arranged in a deliberate visual order and often partially or fully occlude one another. 
This layered structure is central to how designs are created, interpreted, edited, and reused. 
Consequently, access to layer-level representations is essential for understanding and modifying graphic designs~\cite{zhu2026infodet, lin2026mildedit}. 
In practice, however, graphic designs are often distributed as flat raster images (\eg, JPEG or PNG) for compatibility, which removes explicit layer information and makes subsequent editing difficult. 
Recovering the underlying layers from a flat image is therefore a key step for downstream applications~\cite{yin2025qwen}. 
A fundamental task in this process is graphic design detection, which aims to identify the locations, categories, and compositional order of design elements~\cite{suzuki2025layerd}.

Recent advances in visual recognition and object detection~\cite{zou2023object,xiao2025towards,liu2017towards} have led to increasingly capable models that accurately localize and categorize a broad range of objects.
However, these models are primarily designed for natural images~\cite{jiang2025rexomni,ren2024dino} or text-dominant documents~\cite{huang2022layoutlmv3,sun2025pp,pfitzmann2022doclaynet}, which introduces two key limitations when applied to graphic design detection.
First, objects in these two domains generally lack an intrinsic compositional order, so existing models represent them as an unordered set and can not predict the compositional order of the elements.
Without this order, the recovered elements can not be faithfully recomposed into the original design.
Second, because annotating large-scale occluded regions is challenging, existing models are usually trained to predict only visible regions.
As a result, the hidden parts of partially occluded elements remain unrecovered.
These limitations call for a graphic design detection model that can infer compositional order while also learning to recover occluded regions from large-scale annotated data.

To fill this gap, we present \textbf{D}etect \textbf{A}nything in Graphic \textbf{D}esign (\model{}), a model for graphic design detection.
Building upon autoregressive detection with vision-language models (VLMs)~\cite{jiang2025rexomni, chen2021pix2seq}, \model{} formulates graphic design detection as compositional deconstruction and predicts elements in compositional order.
Each element is specified by its category and an amodal bounding box that captures its full spatial extent, including the regions occluded by the elements above it.
With compositional order, \model{} uses lower-layer elements as context when detecting a higher-layer one, aligning with the layer-by-layer process by which designers construct graphic designs.
Based on this formulation, we propose \textbf{Ele}ment \textbf{R}elative \textbf{P}olicy \textbf{O}ptimization (\alg{}), which extends Group Relative Policy Optimization (GRPO)~\cite{shao2024deepseekmath} from sequence-level supervision to element-level optimization.
At the core of \alg{} is the martingale difference decomposition~\citep{williams1991probability}, which decomposes a sequence-level reward into unbiased, history-conditioned element-level rewards.
Estimating this reward requires no additional reward models or extra sampling passes, making it a simple drop-in replacement for GRPO and applicable to any VLM-based detection model.
To support the training and evaluation of \model{}, we construct a large-scale dataset of $10$ million graphic designs, \dataset{}.
The dataset is anchored by a corpus of $5$ million designer-created works, from which we extract ground-truth amodal bounding boxes and compositional order using their original layered representations.
To further improve the diversity of occluded elements and styles, we augment this corpus with an additional $5$ million designs, including those constructed via element replacement, as well as designer-created and AI-generated flat designs.

We validate the effectiveness of \model{} and \alg{} through comprehensive experiments.
\model{} achieves better detection performance than both proprietary and open-source baselines on our dataset, and these gains transfer to existing graphic design detection benchmarks (Fig.~\ref{fig:qualitative}).
Human evaluation further shows that \model{} achieves human-level accuracy in predicting the occluded regions of elements.
In addition, \alg{} consistently improves over GRPO and other fine-grained reinforcement learning (RL) methods, such as PRIME~\cite{cui2025process} and  VinePPO~\cite{kazemnejad2025vineppo}, and generalizes across detection models and a wide range of detection benchmarks.

The main contributions of this work are threefold:
\begin{itemize}[left=3mm,itemsep=0mm,topsep=0mm]
    \item We develop a graphic design detection model, \model{}, that predicts the amodal bounding boxes of elements in their compositional order.
   \item We propose an optimization method, \alg{}, that extends GRPO to element-level supervision by decomposing sequence-level rewards into a martingale difference sequence. 
   \item We release a subset of \dataset{} containing $100,000$ designs for image-to-layer decomposition, each with original layer representations and diverse element styles.
\end{itemize}

\section{Related Work}
\label{sec:related}

\paragraph{Autoregressive Object Detection}
Object detection has evolved from set-based detectors~\cite{carion2020detr,zhang2023dino,zong2023detrs,liu2024grounding,jia2023hdetr} to autoregressive detectors with fixed categories~\cite{chen2021pix2seq,chen2022pix2seqv2,wang2022ofa,lu2023unifiedio}, and more recently to promptable VLM-based models for open-vocabulary detection~\cite{wang2023visionllm,xiao2024florence,jiang2025rexomni,shen2025vlm} and grounding~\cite{peng2023kosmos2,you2024ferret}. 
These advances have also influenced structured document domains, where layout-aware document understanding models~\cite{xu2020layoutlm,xu2021layoutlmv2,huang2022layoutlmv3,liao2025doclayllm,liu2026parl}, specialized layout detectors~\cite{zhao2024doclayoutyolo,sun2025pp}, and benchmarks~\cite{zhong2019publaynet,pfitzmann2022doclaynet,li2020docbank,ouyang2025omnidocbench,heo2025led} have supported structured document analysis.
However, existing detectors represent objects as an unordered set, and thus cannot recover the compositional order and occluded regions required for layer reconstruction.
Closing this gap typically requires learning this information from human annotations~\cite{zhu2017semantic,qi2019kins}, which are difficult to scale and can be imprecise.
In contrast, \model{} learns from exact ground truth extracted at scale from original layered representations, enabling compositional-order prediction and occluded-region recovery.

\paragraph{Reward Design for RL Post-Training}
A common RL algorithm is GRPO~\cite{shao2024deepseekmath}, which assigns a sequence-level reward uniformly to all tokens. 
Existing extensions to finer-grained supervision can be grouped into two categories: policy-signal-based methods and re-rollout-based methods.
Policy-signal-based methods derive fine-grained supervision from the current rollout, either by directly using policy statistics~\cite{tan2025gtpo,parthasarathi2025grpo}, or by learning reward models over intermediate states~\cite{cui2025process,liang2025spodiffusion,savani2026stepwiseflowgrpo}.
Re-rollout-based methods produce fine-grained supervision by sampling additional rollouts and comparing them ~\cite{kazemnejad2025vineppo,guo2026segment,ahmadian2024rloo,lai2024stepdpo}.
These methods are designed for settings where intermediate tokens can not be directly evaluated.
Therefore, they rely on proxy supervision that is not directly aligned with the final reward.
In detection tasks, however, each predicted element contributes measurably to the final reward. 
\alg{} exploits this property to derive element-level rewards via martingale difference decomposition, without requiring additional reward models or re-rollouts. 
To our knowledge, \alg{} is the first method that leverages the inherent decomposability of detection rewards.

\section{Method}
\label{sec:method}

We formulate graphic design detection as compositional deconstruction, introduce \alg{} for element-level optimization, and construct \dataset{}, a 10-million dataset with layer annotations.

\subsection{Task Formulation}

Predicting the compositional order from an image is challenging because occlusion relationships are implicit, and the model must infer which element lies above the other when they overlap.
We address this by formulating graphic design detection as compositional deconstruction, where elements are generated sequentially in compositional order.
Given an input image $I$, the model predicts an element sequence $(e_1,\dots,e_m)$ with policy $e_k\sim \pi_\theta(\cdot|e_{<k}, I)$, using previously predicted lower-layer elements $e_{<k}$ as context for the higher-layer one $e_k$.
Sec.~\ref{subsubsec:EleRPO_ablation} shows that this formulation performs better than alternative element orderings.

\subsection{Element Relative Policy Optimization}

GRPO optimizes a sequence-level reward that measures the overall quality of predicted elements~\cite{jiang2025rexomni, shen2025vlm}.
In our implementation, we use the IoU-weighted $F_1$~\cite{jiang2025rexomni}, a commonly used reward function that reflects localization accuracy, precision, and recall.
It is computed by matching predicted elements to ground truth via spatial overlap, assigning each prediction a score $s_i$ equal to its IoU if the category is correct, and 0 otherwise, and then calculating the $F_1$ score from these scores.
GRPO assigns the same sequence-level advantage to all tokens, thereby ignoring each element’s contribution to the final detection quality.
As a result, an inaccurately predicted element in a high-reward rollout still receives a positive advantage, inadvertently reinforcing incorrect predictions.
To address this limitation, we first decompose the sequence-level reward into element-level rewards via martingale difference decomposition.
We then estimate the element-level rewards from a group of sampled rollouts and convert them into the advantages for policy updates.

\subsubsection{Element-Level Reward via Martingale Difference Decomposition}
\label{subsubsec:martingale}

Ideal element-level rewards $\{r_k\}_{k=1}^m$ should satisfy three properties:
\begin{enumerate}[leftmargin=2em, itemindent=-3pt]
    \item \textbf{Completeness}: $\sum_{k=1}^m r_k = F_1-\mathbb{E}[F_1|I]$. The rewards should sum to the total change of the final $F_1$ score over its expectation before any element is generated.
    \item \textbf{Causality}: $r_k$ depends only on $e_{\leq k}$. Since elements are generated autoregressively, the reward for $e_k$ should depend only on the lower-layer elements, without access to higher-layer elements.
    \item \textbf{Unbiasedness}: $\mathbb{E}[r_k| e_{<k},I]=0$. The reward should isolate the contribution of $e_k$ itself and not be biased by lower-layer elements.
\end{enumerate}

These properties uniquely define the element-level rewards as the martingale differences~\cite{williams1991probability} of $F_1$, which captures how the conditional expected $F_1$ evolves as elements are generated:
\begin{equation}
r_k = \mathbb{E}[F_1 | e_{\leq k}, I] - \mathbb{E}[F_1 | e_{<k}, I].
\label{eq:martingale}
\end{equation}
The conditional expectation $\mathbb{E}[F_1 | e_{\leq k}, I]$ decomposes into $F_1(k)$ and the expected remaining improvement $\mathbb{E}[F_1-F_1(k) | e_{\leq k}, I]$.
Here, $F_1(k)$ is the score computed by comparing the lower-layer elements $e_{\leq k}$ with the ground-truth elements. 
Substituting this decomposition into Eq.~(\ref{eq:martingale}) gives:
\begin{equation}
r_k = \Delta F_1(k) - (\mathbb{E}[F_1{-}F_1(k{-}1)|e_{<k}, I] - \mathbb{E}[F_1{-}F_1(k)|e_{\leq k}, I]),
\label{eq:martingale_final}
\end{equation}
where $\Delta F_1(k) {=} F_1(k){-}F_1(k{-}1)$ is the change in $F_1$ score after adding $e_k$ to the lower-layer elements $e_{<k}$.
Detailed proofs are provided in Appendix~\ref{app:theory:martingale}.

\begin{figure}[!b]
\centering
    \centering
    \includegraphics[width=1\linewidth]{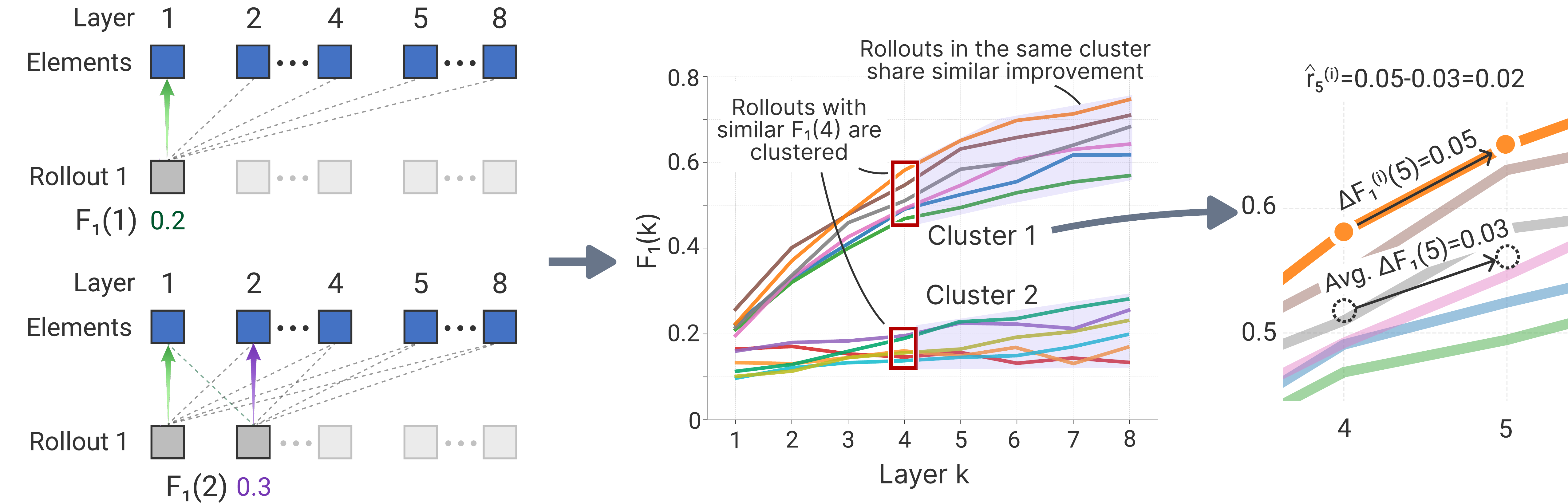}
    \caption{The estimation of element-level reward in \alg.}
\label{fig:method}
\end{figure}

\subsubsection{Advantage Estimation from Sampled Rollouts}
\label{subsubsec:elerpo_estimate}

While Eq.~(\ref{eq:martingale_final}) defines the ideal element-level reward, exactly computing the expected improvements, $\mathbb{E}[F_1{-}F_1(k{-}1)| e_{<k}, I]$ and 
$\mathbb{E}[F_1{-}F_1(k)| e_{\leq k}, I]$, is intractable. 
As shown in Fig.~\ref{fig:method}, we approximate them using rollouts sampled for each image, and then estimate the element-level rewards.

A key challenge is that different rollouts have different lower-layer elements $e_{< k}$, making their expected improvement not directly comparable.
Under the compositional order, however, rollouts naturally form clusters based on the detection quality of lower-layer elements, and rollouts within the same cluster exhibit similar expected improvement in $F_1$.
Inspired by Pavlenko~\etal~\cite{pavlenko2026blockwise}, we group rollouts with similar $F_1(k{-}1)$ and approximate $\mathbb{E}[F_1{-}F_1(k{-}1)|e_{< k}, I] \approx \mathbb{E}[F_1{-}F_1(k{-}1)|F_1(k{-}1), I]$.
In practice, we partition the rollouts into $N_c$ clusters using k-means on $F_1(k{-}1)$, where $N_c$ is selected based on the silhouette score.
This approximation avoids additional re-rollouts, yet achieves a Pearson correlation of $0.81$ with an expensive but accurate re-rollout-based estimator.
We use the same clusters to estimate both expected improvements in Eq.~(\ref{eq:martingale_final}).
This makes their difference a paired comparison across the same rollouts, avoiding the bias introduced by using different clusters.
Let $\mathcal{C}_{k}(i)$ denote the cluster containing rollout $i$.
The element-level reward is then estimated as:
\begin{equation}
\hat{r}_k^{(i)} = \Delta F_1^{(i)}(k) - \frac{1}{|\mathcal{C}_{k}(i)|} \sum\nolimits_{j \in \mathcal{C}_{k}(i)} \Delta F_1^{(j)}(k).
\label{eq:method2}
\end{equation}
Detailed derivations, the justification for using the same clusters, and the details on the empirical validation are provided in Appendix~\ref{app:theory}.
The magnitude of $\hat{r}_k^{(i)}$ scales as $\mathcal{O}(1/(k+n))$, where $n$ is the number of ground-truth elements (see Appendix~\ref{app:theory:magnitude} for the proof).
Without correction, the lower-layer elements would dominate policy updates because their rewards have larger magnitudes.
To address this, we normalize the reward by the layer-wise standard deviation.
Let $\mathcal{G}_k$ denote the set of rollouts that contain at least $k$ elements. 
The element-level advantage is:
\begin{equation}
\hat{A}_k^{(i)} = \hat{r}_k^{(i)}\big/\sqrt{\frac{1}{|\mathcal{G}_k|}\sum\nolimits_{j \in \mathcal{G}_k} (\hat{r}_k^{(j)})^2 + \epsilon}.
\label{eq:normalized-advantage}
\end{equation}
While $\hat{A}_k^{(i)}$ provides fine-grained supervision, it is estimated from a finite group of rollouts and can be noisy.
To improve stability, we combine it with the sequence-level GRPO advantage $\hat{A}^{(i)}$:
\begin{equation}
\resizebox{0.92\linewidth}{!}{$
\mathcal{J}_{\text{\alg}}(\theta) = \frac{1}{G}\sum_{i=1}^{G} \frac{1}{|o^{(i)}|}\sum_{t=1}^{|o^{(i)}|} \left[\min\!\Big(\rho_{i,t} \tilde{A}_{\phi(i,t)}^{(i)},\; \text{clip}(\rho_{i,t}, 1-\epsilon, 1+\epsilon) \tilde{A}_{\phi(i,t)}^{(i)}\Big) - \beta \mathbb{D}_{\text{KL}}[\pi_\theta \Vert \pi_{\text{ref}}]\right]
$}
\label{eq:alg-objective}
\end{equation}
where $\tilde{A}_{\phi(i,t)}^{(i)} = \tfrac{1}{2}\big(\hat{A}_{\phi(i,t)}^{(i)} + \hat{A}^{(i)}\big)$, and $\phi(i,t)$ maps token $o_t^{(i)}$ to its corresponding element.

\subsection{Dataset Construction}

We construct the dataset, \dataset{}, in two stages: data collection and data selection. 

\paragraph{Data collection}
The dataset is anchored by an internal corpus of $5,367,586$ designer-created works.
These works preserve their original layered representations, so we automatically extract ground-truth amodal bounding boxes and compositional orders. 
Such annotations would be infeasible to obtain manually at this scale.
To further improve diversity, we augment this corpus with three sources:
1) $2,075,896$ designs created by replacing elements with stylistically and semantically similar ones retrieved from a pool of 12 million elements;
2) $127,931$ designs generated using Ideogram~\citep{ideogram} and annotated by Gemini-3-Pro~\citep{gemini3pro}; and
3) $2,845,242$ internal flat designer-created designs annotated by Gemini-3-Pro.
Together, these sources result in a dataset of $10,416,655$ designs.
For public data release, we sample $100,000$ designs with replaced elements, each annotated with the original layer representations.

\paragraph{Data selection}
Despite the availability of layer annotations, the model still faces granularity ambiguity: semantically independent elements may be merged into one (\eg, a cluster of decorative stars), while a single coherent element may be split into multiple parts (\eg, a logo decomposed into several decorative shapes).
We address this issue with a two-step filtering process.
First, to improve the separation of small, independent elements, we train a preliminary detector and retain designs that achieve low recall for such elements.
These designs provide useful supervision for learning to separate elements that are easily merged.
Second, to discourage over-splitting, we filter out designs containing many small, mutually overlapping elements clustered in a small region, as these often correspond to parts of a single coherent element.
This process yields $24,195$ high-quality designs, which we split into $19,195$ designs for RL training and $5,000$ for testing.

More details on data collection and selection are provided in Appendix~\ref{app:dataset}.

\subsection{Implementation Details}
\label{sec:implementation}

We build \model{} upon Qwen3-VL-2B~\citep{bai2025qwen3}.
Each element is represented by a bounding box with coordinates normalized to $[0, 1000]$, and a category label indicating whether the element is text or visual.
We train \model{} with LoRA~\citep{hu2022lora} using rank $32$ across all linear layers.
We first perform supervised fine-tuning (SFT) for $650,000$ steps with a learning rate of $1e{-}4$ and a batch size of 64.
This stage uses all data in \dataset{} except the RL training and test splits.
This is followed by RL using \alg{} for $5,000$ steps, with $16$ rollouts per image and a learning rate of $5e{-}7$.
Following Liu~\etal~\citep{liu2026gdpo}, we compute separate IoU-weighted $F_1$ rewards for text and visual elements, and use their average as the final reward.
More training details are provided in Appendix~\ref{app:impl}.

\section{Experiments}
\label{sec:exp}

\subsection{Performance of \model}

To demonstrate the effectiveness of \model{}, we compare it with state-of-the-art models, evaluate its amodal detection quality through human evaluation, and apply it to layer decomposition.

\begin{table}[!b]
\centering
\vspace{-4mm}
\caption{Comparison of \model{} with state-of-the-art baselines on graphic design benchmarks.}
\label{tab:generalization}
\tablestyle{2pt}{1.2}
\resizebox{\linewidth}{!}{%
\begin{tabular}{l|cc|c|c|cc|c|c}
    \shline
    & \multicolumn{4}{c|}{\textit{\dataset{} test set}} & \multicolumn{4}{c}{\textit{DMLB}} \\
    \hline
    Method & $F_1$@[.5:.95]$\uparrow$ & IoU-weighted $F_1$$\uparrow$ & Occluded R@[.5:.95]$\uparrow$ & POA$\uparrow$ & $F_1$@[.5:.95]$\uparrow$ & IoU-weighted $F_1$$\uparrow$ & Occluded R@[.5:.95]$\uparrow$ & POA$\uparrow$ \\
    \hline
    Rex-Omni           & 31.24 & 38.12 & 15.41 & - & 17.79 & 19.19 & 13.25 & - \\
    Gemini-3.1-Pro     & 32.58 & 39.95 & 25.15 & 16.56 & 44.17 & 52.60 & 42.68 & 22.57 \\
    GPT-5.5            & 40.89 & 49.46 & 34.26 & 25.42 & 61.05 & 70.24 & 61.35 & 34.17 \\
    \rowcolor{blue!10} \model{} (Ours) & \textbf{72.57} & \textbf{79.65} & \textbf{65.35} & \textbf{50.18} & \textbf{79.52} & \textbf{84.42} & \textbf{71.65} & \textbf{46.97} \\
    \shline
    & \multicolumn{4}{c|}{\textit{Crello}} & \multicolumn{4}{c}{\textit{PrismLayersPro}} \\
    \hline
    Method & $F_1$@[.5:.95]$\uparrow$ & IoU-weighted $F_1$$\uparrow$ & Occluded R@[.5:.95]$\uparrow$ & POA$\uparrow$ & $F_1$@[.5:.95]$\uparrow$ & IoU-weighted $F_1$$\uparrow$ & Occluded R@[.5:.95]$\uparrow$ & POA$\uparrow$ \\
    \hline
    Rex-Omni           & 26.04 & 33.97 & 22.31 & - & 35.14 & 38.29 & 25.42 & - \\
    Gemini-3.1-Pro     & 34.32 & 42.20 & 36.91 & 18.77 & 40.75 & 48.54 & 37.71 & 22.32 \\
    GPT-5.5            & 45.86 & 56.53 & 52.26 & 29.31 & 61.41 & 73.66 & 60.53 & 37.29 \\
    \rowcolor{blue!10} \model{} (Ours) & \textbf{57.75} & \textbf{68.31} & \textbf{58.03} & \textbf{34.89} & \textbf{73.21} & \textbf{78.44} & \textbf{72.67} & \textbf{43.87} \\
    \shline
    & \multicolumn{4}{c|}{\textit{LICA}} & \multicolumn{4}{c}{\textit{ChartGalaxy}} \\
    \hline
    Method & $F_1$@[.5:.95]$\uparrow$ & IoU-weighted $F_1$$\uparrow$ & Occluded R@[.5:.95]$\uparrow$ & POA$\uparrow$ & $F_1$@[.5:.95]$\uparrow$ & IoU-weighted $F_1$$\uparrow$ & Occluded R@[.5:.95]$\uparrow$ & POA$\uparrow$ \\
    \hline
    Rex-Omni           & 28.51 & 35.70 & 16.37 & - & 6.31 & 8.24 & 3.16 & - \\
    Gemini-3.1-Pro     & 39.42 & 47.90 & 35.53 & 22.17 & 12.29 & 17.95 & 9.27 & 6.92 \\
    GPT-5.5            & 45.92 & 54.95 & 45.10 & 29.53 & 6.85 & 9.42 & 2.98 & 1.12 \\
    \rowcolor{blue!10} \model{} (Ours) & \textbf{71.02} & \textbf{78.89} & \textbf{62.67} & \textbf{46.96} & \textbf{32.97} & \textbf{57.71} & \textbf{35.76} & \textbf{27.71} \\
    \shline
\end{tabular}%
}
\end{table}

\subsubsection{Comparison with State-of-the-Art}
\label{subsec:main_comparison}

\paragraph{Datasets}
We evaluate \model{} on \dataset{} test set and five graphic design benchmarks: Design-Multi-Layer-Bench (DMLB)~\cite{pu2025art}, Crello~\cite{yamaguchi2021canvasvae}, PrismLayersPro~\cite{chen2025prismlayers}, LICA~\cite{hirsch2026lica}, and ChartGalaxy~\cite{li2025chartgalaxy}.
The \dataset{} test set is held out from training, and a benchmark-wise overlap check finds no duplicates between \dataset{} and the five external benchmarks.

\paragraph{Evaluation metrics}
We evaluate the performance from four aspects:
1) overall detection quality, measured by the $F_1$ score averaged over IoU thresholds from $0.5$ to $0.95$ ($F_1$@[.5:.95]) and the IoU-weighted $F_1$;
2) amodal detection quality, measured by recall on elements with at least 5\% occluded area, averaged over IoU thresholds (Occluded R@[.5:.95]);
3) ordering accuracy, measured by the pairwise ordering accuracy (POA)~\cite{zhu2017semantic}, defined as the proportion of correctly ordered overlapping element pairs, where a pair is correct if both elements are matched to ground-truth elements and their predicted order matches the ground-truth order; and
4) inference efficiency, measured by the average inference time per design.

\paragraph{Baselines}
We compare \model{} with a set of representative proprietary and open-source models.
For proprietary models, we use GPT-4o~\cite{gpt4o}, GPT-5~\cite{gpt5}, GPT-5.4~\cite{gpt5_4}, GPT-5.5~\cite{gpt5_5}, and Gemini-3.1-Pro~\cite{gemini3_1pro}.
For open-source models, we evaluate GLIP-L~\cite{li2022grounded}, Detic-B~\cite{zhou2022detecting}, Grounding DINO-B~\cite{liu2024grounding}, Florence2-L~\cite{xiao2024florence}, Qwen3-VL~\cite{bai2025qwen3}, InternVL3.5~\cite{wang2025internvl3}, and Rex-Omni~\cite{jiang2025rexomni}.

Dataset statistics, data overlap checks, and baseline implementations are detailed in Appendix~\ref{app:eval}.

\paragraph{Results}
We evaluate the models over $3$ runs and report their mean results, except for proprietary models, which are evaluated once due to the cost of API calls.
Table~\ref{tab:generalization} compares \model{} with state-of-the-art proprietary baselines (Gemini-3.1-Pro and GPT-5.5) and open-source baseline (Rex-Omni).
The full results, including the variance across runs and inference efficiency, are provided in Appendix~\ref{app:eval_results}.
\model{} achieves the best performance across all six datasets, improving overall detection quality, amodal detection quality, and ordering accuracy.
It is also more efficient than proprietary baselines and comparable to open-source ones.
As shown in Fig.~\ref{fig:qualitative}(a), \model{} performs robustly across diverse designs, detecting elements missed by GPT-5.5 and recovering their occluded regions.
More examples are provided in Appendix~\ref{app:eval_results}.

\subsubsection{Human Evaluation on Amodal Detection}

\begin{wrapfigure}{r}{0.44\linewidth}
    \centering
    \includegraphics[width=\linewidth]{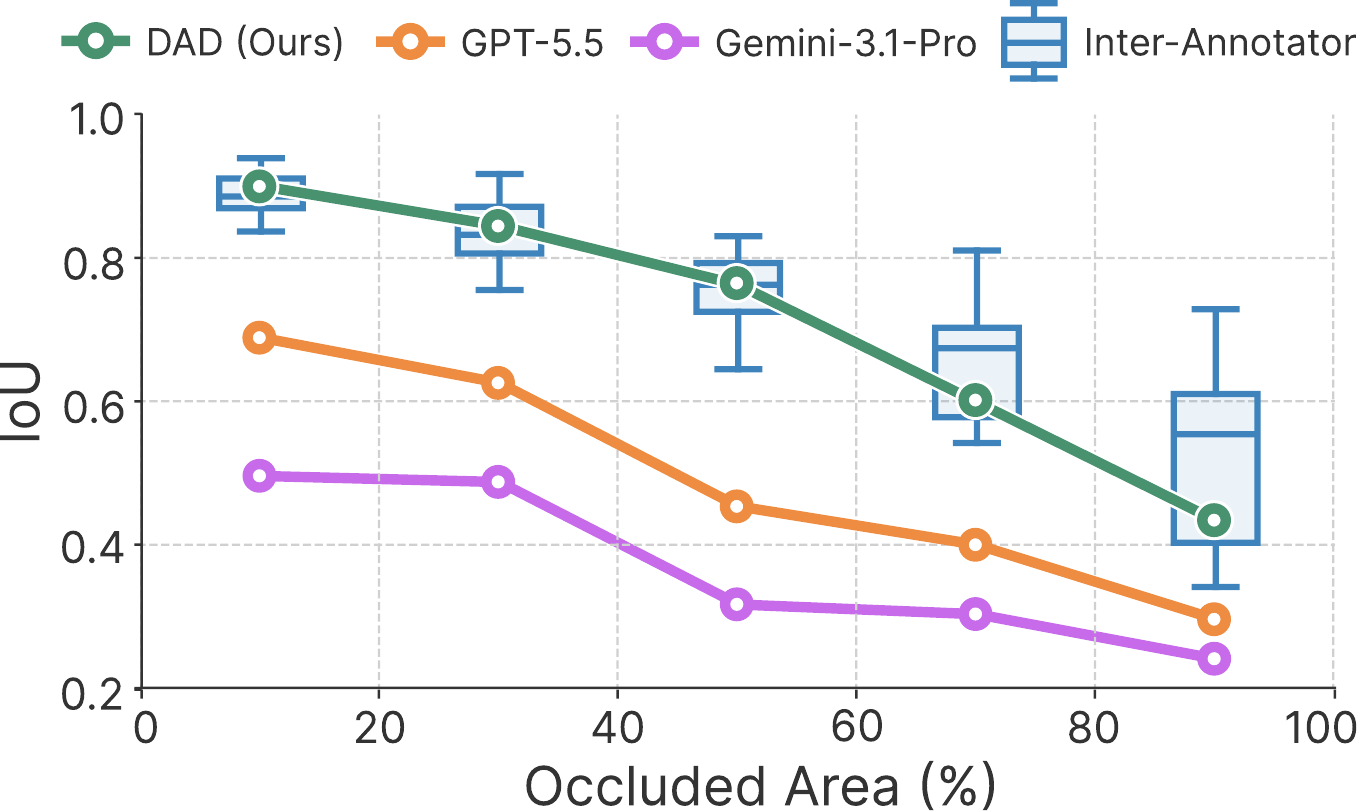}
    \caption{Results of human evaluation on amodal detection.}
    \label{fig:human_eval}
    \vspace{-4mm}
\end{wrapfigure}
\paragraph{Experimental setup}
To evaluate \model{}'s amodal detection quality, we compare its predictions with human annotations on $500$ occluded elements sampled from the \dataset{} test set.
These elements are evenly distributed across five bins, ranging from $0$\%-$20$\% to $80$\%-$100$\% occluded area.
We recruit $16$ annotators with at least one year of graphic design experience, and divide them into $4$ groups. 
Each group contains $4$ annotators and annotates the same $125$ elements.
Annotators are shown only the visible region of each element and asked to label its amodal bounding box using a web-based annotation tool.
The annotation interface and compensation details are provided in Appendix~\ref{app:human}.
For each element, we obtain its predictions from \model{}, GPT-5.5, and Gemini-3.1-Pro by selecting the one with the highest IoU.
We then compute the average IoU between model predictions and human annotations.
As a reference for human-level performance, we also compute inter-annotator IoU within each group.

\paragraph{Results}
Fig.~\ref{fig:human_eval} compares the average IoU between each model and human annotators against the distribution of inter-annotator IoUs.
GPT-5.5 and Gemini-3.1-Pro consistently fall below the inter-annotator IoU range across all occlusion levels.
In contrast, \model{} stays within this range, indicating that it achieves human-level amodal detection quality.

\begin{table}[!b]
\centering
\vspace{-4mm}
\caption{Evaluation results on image-to-layer decomposition across graphic design benchmarks.}
\label{tab:layer_decomp}
\tablestyle{2pt}{1.2}
\resizebox{\linewidth}{!}{
\begin{tabular}{l|cc|cc|cc|cc|cc|cc}
    \shline
    & \multicolumn{6}{c|}{\textit{\dataset{} test set}} & \multicolumn{6}{c}{\textit{DMLB}} \\
    \hline
    Method & PSNR$\uparrow$ & SSIM$\uparrow$ & RGB L1$\downarrow$ & Alpha SIoU$\uparrow$ & Occ-PSNR$\uparrow$ & Occ-SSIM$\uparrow$ & PSNR$\uparrow$ & SSIM$\uparrow$ & RGB L1$\downarrow$ & Alpha SIoU$\uparrow$ & Occ-PSNR$\uparrow$ & Occ-SSIM$\uparrow$ \\
    \hline
    GT Box             & 30.23 & 0.955 & 0.188 & 0.592 & 15.42 & 0.468 & 29.28 & 0.947 & 0.213 & 0.553 & 19.79 & 0.572 \\
    \hline
    Gemini-3.1-Pro     & 26.59 & 0.933 & 0.252 & 0.369 & 5.76 & 0.144 & 26.72 & 0.935 & 0.244 & 0.459 & 7.33 & 0.213 \\
    GPT-5.5            & 28.62 & 0.948 & 0.242 & 0.436 & 7.48 & 0.204 & 29.92 & 0.956 & 0.228 & 0.530 & 10.80 & 0.312 \\
    \rowcolor{blue!10} \model{} (Ours) & \textbf{29.83} & \textbf{0.953} & \textbf{0.223} & \textbf{0.511} & \textbf{12.20} & \textbf{0.357} & \textbf{30.10} & \textbf{0.957} & \textbf{0.210} & \textbf{0.571} & \textbf{14.23} & \textbf{0.383} \\
    \shline
    & \multicolumn{6}{c|}{\textit{PrismLayersPro}} & \multicolumn{6}{c}{\textit{Crello}} \\
    \hline
    Method & PSNR$\uparrow$ & SSIM$\uparrow$ & RGB L1$\downarrow$ & Alpha SIoU$\uparrow$ & Occ-PSNR$\uparrow$ & Occ-SSIM$\uparrow$ & PSNR$\uparrow$ & SSIM$\uparrow$ & RGB L1$\downarrow$ & Alpha SIoU$\uparrow$ & Occ-PSNR$\uparrow$ & Occ-SSIM$\uparrow$ \\
    \hline
    GT Box             & 31.02 & 0.950 & 0.178 & 0.499 & 12.04 & 0.417 & 27.44 & 0.913 & 0.125 & 0.783 & 13.87 & 0.415 \\
    \hline
    Gemini-3.1-Pro     & 28.69 & 0.943 & 0.201 & 0.371 & 6.49 & 0.215 & 25.14 & 0.907 & 0.191 & 0.581 & 4.29 & 0.137 \\
    GPT-5.5            & 30.84 & 0.957 & 0.190 & 0.477 & 8.80 & 0.293 & 23.95 & 0.875 & 0.173 & 0.604 & 4.94 & 0.147 \\
    \rowcolor{blue!10} \model{} (Ours) & \textbf{31.40} & \textbf{0.959} & \textbf{0.186} & \textbf{0.482} & \textbf{9.72} & \textbf{0.325} & \textbf{27.83} & \textbf{0.930} & \textbf{0.162} & \textbf{0.642} & \textbf{5.74} & \textbf{0.180} \\
    \shline
    & \multicolumn{6}{c|}{\textit{LICA}} & \multicolumn{6}{c}{\textit{ChartGalaxy}} \\
    \hline
    Method & PSNR$\uparrow$ & SSIM$\uparrow$ & RGB L1$\downarrow$ & Alpha SIoU$\uparrow$ & Occ-PSNR$\uparrow$ & Occ-SSIM$\uparrow$ & PSNR$\uparrow$ & SSIM$\uparrow$ & RGB L1$\downarrow$ & Alpha SIoU$\uparrow$ & Occ-PSNR$\uparrow$ & Occ-SSIM$\uparrow$ \\
    \hline
    GT Box             & 29.98 & 0.952 & 0.195 & 0.637 & 14.97 & 0.443 & 30.80 & 0.974 & 0.235 & 0.426 & 18.25 & 0.506 \\
    \hline
    Gemini-3.1-Pro     & 26.46 & 0.935 & 0.243 & 0.452 & 5.87 & 0.185 & 29.59 & 0.966 & 0.278 & 0.225 & 4.05 & 0.100 \\
    GPT-5.5            & 29.20 & 0.955 & 0.224 & 0.534 & 8.06 & 0.255 & 29.27 & 0.968 & 0.272 & 0.270 & 4.11 & 0.093 \\
    \rowcolor{blue!10} \model{} (Ours) & \textbf{29.81} & \textbf{0.956} & \textbf{0.222} & \textbf{0.550} & \textbf{9.94} & \textbf{0.324} & \textbf{30.10} & \textbf{0.972} & \textbf{0.270} & \textbf{0.328} & \textbf{9.79} & \textbf{0.270} \\
    \shline
\end{tabular}
}
\end{table}

\subsubsection{Application to Image-to-Layer Decomposition}

\paragraph{Experimental setup}
We evaluate \model{} on image-to-layer decomposition using the MRT model~\cite{tang2026mrt}, which recovers the underlying layers of a flat image based on input bounding boxes of elements. 
We compare \model{} with GPT-5.5 and Gemini-3.1-Pro by using their predicted bounding boxes as input to MRT, and evaluate the resulting layers on the same six datasets as in Section~\ref{subsec:main_comparison}.
As a reference, we also evaluate MRT with ground-truth boxes as input.
We evaluate the performance from three aspects:
1) image-level recovery quality, measured by the PSNR and SSIM of the image recomposed from the predicted layers;
2) layer-level recovery quality, following LayerD~\cite{suzuki2025layerd}, measured by RGB L1 (alpha-weighted L1 distance on RGB channels) and Alpha soft IoU (soft IoU on alpha channels) between predicted and ground-truth layers, allowing no adjacent-layer merges to jointly evaluate layer content and granularity; and
3) occlusion recovery quality, measured by the PSNR and SSIM computed on occluded pixels in the ground-truth layers, where predicted layers are matched to ground-truth layers using the Hungarian algorithm, and scores are averaged over occluded ground-truth layers, with unmatched layers assigned a score of $0$.

\paragraph{Results}
As shown in Table~\ref{tab:layer_decomp}, \model{} consistently outperforms GPT-5.5 and Gemini-3.1-Pro across all datasets, improving image-level, layer-level, and occlusion recovery quality.
On image-level PSNR and SSIM, \model{} matches the results achieved with ground-truth boxes, indicating that its predicted boxes are nearly as informative for recovering visible regions.
As shown in Fig.~\ref{fig:qualitative}(b), these predictions enable faithful layer recovery, including occluded regions missed by GPT-5.5.

\subsection{Effectiveness of \alg}

To demonstrate the effectiveness of \alg{}, we compare it with existing RL methods, conduct ablation studies on its key components, and evaluate its generalizability across models and benchmarks.

\begin{table}[!b]
\vspace{-4mm}
\centering
\caption{Comparison of \alg{} with existing RL methods on \dataset{}.}
\label{tab:elrpo_comparison}
\tablestyle{4pt}{1.3}
\resizebox{0.7\linewidth}{!}{
\begin{tabular}{c|c|ccc|cc|c}
    \shline
    SFT& GRPO & GRPO-$\lambda$ & GTPO & PRIME & VinePPO & SPO & \cellcolor{blue!10}\alg{} (Ours) \\
    \hline
    54.8 & 74.2 & 73.0 & 74.1 & 73.3 & 58.0 & 57.9 & \cellcolor{blue!10}\textbf{79.7} \\
    \shline
\end{tabular}
}
\end{table}

\subsubsection{Comparison with Existing Methods}
\label{subsubsec:elerpo_comparison}

\paragraph{Experimental setup} 
We compare \alg{} with GRPO and its finer-grained variants, including representative policy-signal-based methods (GRPO-$\lambda$~\cite{parthasarathi2025grpo}, GTPO~\cite{tan2025gtpo}, and PRIME~\cite{cui2025process}), as well as re-rollout-based methods (VinePPO~\cite{kazemnejad2025vineppo} and SPO~\cite{guo2026segment}).
All the methods are applied to \model-SFT, which denotes \model{} after supervised fine-tuning, using the RL configuration in Sec.~\ref{sec:implementation}.
Method-specific hyperparameters, such as the number of Monte Carlo continuations for VinePPO and SPO, follow the original papers. 
Additional details are provided in Appendix~\ref{app:elerpo:hparams}.
We evaluate all methods on \dataset{} test set using the training reward, IoU-weighted $F_1$.

\paragraph{Results}
As shown in Table~\ref{tab:elrpo_comparison}, the policy-signal-based methods do not improve over GRPO, as they are designed for chain-of-thought reasoning and fail to capture each element's contribution in detection.
For example, GTPO assigns higher weights to high-entropy tokens, which are often low-order coordinate digits with limited impact on the final detection result.
The re-rollout-based methods perform worse than GRPO because they are designed for tasks with discrete or binary outcomes, where a small number of rollouts can estimate success rates.
In detection, however, reward changes are continuous and fine-grained, making them difficult to estimate reliably from the same number of rollouts.  
\alg{} addresses this limitation by directly decomposing the sequence-level reward into element-level rewards, outperforming GRPO and all other baselines.

\begin{table}[!t]
\centering
\caption{Ablation studies on key components of \alg{}.}
\label{tab:ablation}
\setlength{\tabcolsep}{4pt}
\renewcommand{\arraystretch}{1.2}
\resizebox{\linewidth}{!}{%
\begin{tabular}[t]{@{}c@{\hspace{20pt}}c@{\hspace{20pt}}c@{}}
\begin{tabular}[t]{@{}c@{}}
(a) Reward causality \\[-4pt]
\begin{tabular}[t]{cc}
    \shline
    IoU & \cellcolor{blue!10} $\Delta F_1$ \\
    \hline
    55.8 & \cellcolor{blue!10}\textbf{79.7} \\
    \shline
\end{tabular}
\end{tabular}
&
\begin{tabular}[t]{@{}c@{}}
(b) Advantage Combination \\[-4pt]
\begin{tabular}[t]{ccc}
    \shline
    Sequence-level & Element-level & \cellcolor{blue!10}Combined \\
    \hline
    74.2 & 78.5 & \cellcolor{blue!10}\textbf{79.7} \\
    \shline
\end{tabular}
\end{tabular}
&
\begin{tabular}[t]{@{}c@{}}
(c) Element Order \\[-4pt]
\begin{tabular}[t]{l|ccc}
    \shline
     & Random & Left-edge & \cellcolor{blue!10}Compositional \\
    \hline
    SFT          & 51.0 & 50.4 & \cellcolor{blue!10}\textbf{53.0} \\
    + \alg{}  & 54.0 & 55.4 & \cellcolor{blue!10}\textbf{61.2} \\
    \shline
\end{tabular}
\end{tabular}
\end{tabular}%
}
\end{table}

\subsubsection{Ablation Studies}
\label{subsubsec:EleRPO_ablation}

We conduct ablation studies on \alg{} from two perspectives:
1) destructive comparisons that validate the causal reward design, mean subtraction, and standard deviation normalization by intentionally violating each of them; and
2) comparisons of alternative choices for prefix clustering, advantage combination, and element order.
The training and evaluation settings are the same as in Sec.~\ref{subsubsec:elerpo_comparison}, except for the comparison on element order, which requires a separate SFT stage as detailed in Appendix~\ref{app:elerpo:order}.
We report the results on reward causality, advantage combination, and element order in Table~\ref{tab:ablation}.
The remaining ablations are provided in Appendix~\ref{app:elerpo_results}.

\paragraph{Reward causality}
We compare our element-level reward with a non-causal IoU-based reward, which replaces $\Delta F_1(k)$ with the IoU between each predicted element and its Hungarian-matched ground truth, or 0 if unmatched.
Table~\ref{tab:ablation}(a) shows that this non-causal reward leads to a significant performance drop.
This result highlights the importance of aligning the element-level reward with the autoregressive nature of VLM-based detection.

\paragraph{Advantage combination}
Table~\ref{tab:ablation}(b) shows that using only the element-level advantage outperforms the sequence-level GRPO advantage, demonstrating the effectiveness of element-level optimization.
Combining both further improves performance, indicating that the sequence-level advantage provides stable supervision that complements the fine-grained element-level advantage.

\paragraph{Element order}
We compare three prediction orders: random, left-edge order that sorts elements by their leftmost x-coordinate, and compositional order.
Table~\ref{tab:ablation}(c) compares their performance under two training settings: SFT alone and SFT followed by \alg{}.
Compositional order improves performance with SFT alone, and the gain becomes large after applying \alg{}.
This indicates that compositional order and element-level rewards are synergistic: element-level rewards are most informative when the prediction order preserves the underlying compositional structure, allowing marginal contributions to reflect the design logic.

\subsubsection{Generalizability of \alg}

\paragraph{Experimental setup}

We evaluate the generalizability of \alg{} across models and benchmarks.
We consider three models: \model-SFT, Rex-Omni-SFT~\cite{jiang2025rexomni}, and Qwen3-VL-2B fine-tuned on ChartGalaxy~\cite{li2025chartgalaxy}.
As shown in Table~\ref{tab:elrpo_generalizability}, we compare \alg{} with GRPO on benchmarks aligned with each model's domain.
These benchmarks cover three categories: 
1) graphic layout benchmarks with underlying layer structure (ChartGalaxy~\cite{li2025chartgalaxy} and Crello~\cite{yamaguchi2021canvasvae});
2) layout benchmarks without explicit layer structure (Rico~\cite{deka2017rico}, PosterLayout~\cite{hsu2023posterlayout}, and DocLayNet~\cite{pfitzmann2022doclaynet}); and 
3) natural image benchmarks (COCO~\cite{lin2014microsoft}, LVIS~\cite{gupta2019lvis}, BDD100K~\cite{yu2020bdd100k}, and Objects~365~\cite{shao2019objects365}).
We perform the same overlap check as in Sec.~\ref{subsec:main_comparison} and find no duplicates between \dataset{} and these benchmarks.
For each model-benchmark pair, we perform RL training and evaluation using the same setting as in Sec.~\ref{subsubsec:elerpo_comparison}.
More details on the benchmarks are provided in Appendix~\ref{app:eval_ele:benchmarks}.

\begin{table}[H]
\vspace{-4mm}
\centering
\caption{Comparison of \alg{} with GRPO across different models and benchmarks.}
\label{tab:elrpo_generalizability}
\tablestyle{3pt}{1.3}
\small

\begin{minipage}{0.7\linewidth}  
\centering

\begin{minipage}{0.88\linewidth}
\centering
(a) DAD-SFT \\[3pt]
\resizebox{\linewidth}{!}{%
\begin{tabular}{l|cc|ccc|c}
    \shline
    Method & ChartGalaxy & Crello & Rico & PosterLayout & DocLayNet & COCO \\
    \hline
    GRPO         & 56.5 & 56.8 & 49.4 & 63.9 & 48.4 & 38.9 \\
    \rowcolor{blue!10} \alg{} (Ours) & \textbf{63.7} & \textbf{58.7} & \textbf{50.4} & \textbf{66.1} & \textbf{52.5} & \textbf{39.4} \\
    \shline
\end{tabular}}
\end{minipage}

\vspace{6pt}

\begin{minipage}[t]{0.635\linewidth}
\centering
(b) Rex-Omni-SFT \\[3pt]
\resizebox{\linewidth}{!}{%
\begin{tabular}{l|cccc}
    \shline
    Method & COCO & LVIS & BDD100K & Objects 365 \\
    \hline
    GRPO         & 58.9 & 53.4 & 49.1 & 50.4 \\
    \rowcolor{blue!10} \alg{} (Ours) & \textbf{59.7} & \textbf{53.9} & \textbf{49.7} & \textbf{51.0} \\
    \shline
\end{tabular}}
\end{minipage}\hfill
\begin{minipage}[t]{0.34\linewidth}
\centering
(c) Fine-tuned Qwen3-VL \\[3pt]
\resizebox{\linewidth}{!}{%
\begin{tabular}{l|c}
    \shline
    Method & ChartGalaxy \\
    \hline
    GRPO         & 63.9 \\
    \rowcolor{blue!10} \alg{} (Ours) & \textbf{66.4} \\
    \shline
\end{tabular}}
\end{minipage}

\end{minipage}  
\end{table}

\paragraph{Results}
Table~\ref{tab:elrpo_generalizability} shows that \alg{} consistently outperforms GRPO across all model-benchmark pairs.
The largest gains are observed on graphic layout benchmarks, where the compositional structure makes element-level rewards particularly effective.
\alg{} also achieves consistent gains on natural image benchmarks where such structure is absent, demonstrating its generalizability.

\section{Conclusion}
\label{sec:conclusion}
In this paper, we introduce \model{}, a graphic design detection model based on compositional deconstruction.
Instead of treating design elements as an unordered set, \model{} models graphic designs as layered compositions, enabling amodal detection that recovers complete element boundaries even under occlusion.
Experiments demonstrate the effectiveness of \model{} for graphic design detection and its utility in downstream applications such as image-to-layer decomposition.
Despite these promising results, several directions remain for future work.
For example, extending this formulation to recover richer compositional structures beyond individual elements could benefit downstream tasks such as design editing, retrieval, and structured design analysis.

% \begin{ack}
% Use unnumbered first level headings for the acknowledgments. All acknowledgments

% Do {\bf not} include this section in the anonymized submission, only in the final paper. You can use the \texttt{ack} environment provided in the style file to automatically hide this section in the anonymized submission.
% \end{ack}

\bibliographystyle{IEEEtranN}
\bibliography{reference}

%%%%%%%%%%%%%%%%%%%%%%%%%%%%%%%%%%%%%%%%%%%%%%%%%%%%%%%%%%%%
\newpage
\appendix

\section{Theoretical Derivations for \alg{}}
\label{app:theory}

\subsection{Proof of the Martingale Difference Decomposition}
\label{app:theory:martingale}

We provide the derivation of Eq.~(\ref{eq:martingale}) and Eq.~(\ref{eq:martingale_final}).
Let $\mathcal F_k = \sigma(I, e_{\leq k})$ be the filtration generated by the input image $I$ and the lower layer elements $e_{\leq k}$. 
Since the final score $F_1$ is a bounded function of the complete rollout, it is integrable. 
Therefore, define

$$
M_k = \mathbb E[F_1 \mid \mathcal F_k].
$$

Then $\{M_k\}_{k=0}^m$ is a martingale with respect to $\{\mathcal F_k\}_{k=0}^m$. 
The martingale difference at step $k$ is

$$
r_k = M_k - M_{k-1}
=
\mathbb E[F_1 \mid \mathcal F_k]
-
\mathbb E[F_1 \mid \mathcal F_{k-1}],
$$

which gives Eq.~(\ref{eq:martingale}) after writing
$\mathcal F_k$ as $(I,e_{\le k})$.

We first verify that the element-level rewards satisfy the three desired properties.

\paragraph{Completeness}
Since $F_1$ is measurable with respect to $\mathcal F_m$, we have $M_m = F_1$. 
Therefore,

$$
\sum_{k=1}^m r_k
=
\sum_{k=1}^m (M_k-M_{k-1})
=
M_m-M_0
=
F_1-\mathbb E[F_1\mid I].
$$

\paragraph{Causality}
The reward $r_k=M_k-M_{k-1}$ is $\mathcal F_k$-measurable, and therefore depends only on $I$ and $e_{\le k}$. It does not require access to higher-layer elements $e_{>k}$.

\paragraph{Unbiasedness}
By the tower property of conditional expectation,

$$
\mathbb E[r_k\mid \mathcal F_{k-1}]
=
\mathbb E[M_k-M_{k-1}\mid \mathcal F_{k-1}]
=
\mathbb E[M_k\mid \mathcal F_{k-1}] - M_{k-1}
=
M_{k-1}-M_{k-1}
=
0.
$$

Equivalently,

$$
\mathbb E[r_k\mid e_{<k},I]=0.
$$

Next, we prove the uniqueness of the rewards satisfying these properties.
Suppose another reward sequence $\{\tilde r_k\}_{k=1}^m$ satisfies the same three properties. 
Define

$$
\tilde S_k = \sum_{\ell=1}^k \tilde r_\ell,
\qquad
\tilde S_0=0.
$$

By causality, $\tilde S_k$ is $\mathcal F_k$-measurable. By unbiasedness,

$$
\mathbb E[\tilde S_k\mid \mathcal F_{k-1}]
=
\tilde S_{k-1}
+
\mathbb E[\tilde r_k\mid \mathcal F_{k-1}]
=
\tilde S_{k-1},
$$

so $\{\tilde S_k\}_{k=0}^m$ is a martingale. 
By completeness,

$$
\tilde S_m = F_1-\mathbb E[F_1\mid I].
$$

Therefore, for any $k$,

$$
\tilde S_k
=
\mathbb E[\tilde S_m\mid \mathcal F_k]
=
\mathbb E[F_1-\mathbb E[F_1\mid I]\mid \mathcal F_k]
=
\mathbb E[F_1\mid \mathcal F_k]-\mathbb E[F_1\mid I].
$$

Taking differences gives

$$
\tilde r_k
=
\tilde S_k-\tilde S_{k-1}
=
\mathbb E[F_1\mid \mathcal F_k]
-
\mathbb E[F_1\mid \mathcal F_{k-1}]
=
r_k.
$$

Hence, the three properties uniquely determine the martingale differences.

We next derive Eq.~(\ref{eq:martingale_final}). Since $F_1(k)$ is computed from the lower-layer elements $e_{\le k}$, it is $\mathcal F_k$-measurable. 
Therefore,

$$
\mathbb E[F_1\mid \mathcal F_k]
=
\mathbb E[F_1(k)+F_1-F_1(k)\mid \mathcal F_k]
=
F_1(k)+\mathbb E[F_1-F_1(k)\mid \mathcal F_k].
$$

Similarly,

$$
\mathbb E[F_1\mid \mathcal F_{k-1}]
=
F_1(k-1)+\mathbb E[F_1-F_1(k-1)\mid \mathcal F_{k-1}].
$$

Substituting these two identities into the martingale difference gives

\begin{align*}
r_k
&=
\mathbb E[F_1\mid \mathcal F_k]
-
\mathbb E[F_1\mid \mathcal F_{k-1}]
\nonumber\\
&=
F_1(k)-F_1(k-1)
+
\mathbb E[F_1-F_1(k)\mid \mathcal F_k]
-
\mathbb E[F_1-F_1(k-1)\mid \mathcal F_{k-1}]
\nonumber\\
&=
\Delta F_1(k)
-
\Big(
\mathbb E[F_1-F_1(k-1)\mid e_{<k},I]
-
\mathbb E[F_1-F_1(k)\mid e_{\le k},I]
\Big),
\end{align*}

where

$$
\Delta F_1(k)=F_1(k)-F_1(k-1).
$$

This proves Eq.~(\ref{eq:martingale_final}).

\subsection{Derivation of the Element-Level Reward Estimator and Justification for Cluster Reuse}
\label{app:theory:advantage}

For rollout $i$, define the residual score after the first $k$ elements as

\begin{equation} \nonumber
R_{k}^{(i)} = F_1^{(i)} - F_1^{(i)}(k).
\end{equation}

Using this notation, Eq.~(\ref{eq:martingale_final}) for rollout $i$ can be written as

\begin{equation} \nonumber
\begin{aligned}
r_k^{(i)}
=
\Delta F_1^{(i)}(k)
-
\Big(
\mathbb E[R_{k-1}\mid e_{<k}^{(i)}, I]
-
\mathbb E[R_k\mid e_{\le k}^{(i)}, I]
\Big).
\end{aligned}
\label{eq:app-residual-form}
\end{equation}

The two conditional residual expectations are expensive to estimate directly.
As discussed in Sec.~\ref{subsubsec:elerpo_estimate}, we use $F_1(k-1)$ as a proxy for the prefix state. 
For each step $k$, rollouts are partitioned according to $F_1(k-1)$, where $\mathcal C_k(i)$ denotes the cluster containing rollout $i$. 
We estimate the difference of residual expectations using a paired comparison within the same cluster:
\begin{equation} \nonumber
\begin{aligned}
\widehat B_k^{(i)}
&=
\frac{1}{|\mathcal C_k(i)|}
\sum_{j\in \mathcal C_k(i)}
\left(R_{k-1}^{(j)}-R_k^{(j)}\right).
\end{aligned}
\label{eq:app-paired-residual}
\end{equation}

For every rollout $j$,
\begin{equation} \nonumber
\begin{aligned}
R_{k-1}^{(j)}-R_k^{(j)}
&=
\big(F_1^{(j)}-F_1^{(j)}(k-1)\big)
-
\big(F_1^{(j)}-F_1^{(j)}(k)\big) \\
&=
F_1^{(j)}(k)-F_1^{(j)}(k-1) \\
&=
\Delta F_1^{(j)}(k).
\end{aligned}
\end{equation}
Therefore,
\begin{equation} \nonumber
\begin{aligned}
\widehat B_k^{(i)}
&=
\frac{1}{|\mathcal C_k(i)|}
\sum_{j\in \mathcal C_k(i)}
\Delta F_1^{(j)}(k).
\end{aligned}
\label{eq:app-cluster-delta-baseline}
\end{equation}
This gives the estimated element-level reward in Eq.~(\ref{eq:method2}):
\begin{equation} \nonumber
\begin{aligned}
\hat r_k^{(i)}
&=
\Delta F_1^{(i)}(k)
-
\widehat B_k^{(i)} \\
&=
\Delta F_1^{(i)}(k)
-
\frac{1}{|\mathcal C_k(i)|}
\sum_{j\in \mathcal C_k(i)}
\Delta F_1^{(j)}(k),
\end{aligned}
\end{equation}

We now justify why the same cluster should be used for the two residual terms. 
Suppose the first residual term is estimated over $\mathcal C_k^-(i)$, while the second residual term is estimated over a different cluster $\mathcal C_k^+(i)$. 
The resulting estimate of the residual difference would be
\begin{equation} \nonumber
\begin{aligned}
\widetilde B_k^{(i)}
&=
\frac{1}{|\mathcal C_k^-(i)|}
\sum_{j\in \mathcal C_k^-(i)}
R_{k-1}^{(j)}
-
\frac{1}{|\mathcal C_k^+(i)|}
\sum_{j\in \mathcal C_k^+(i)}
R_k^{(j)}.
\end{aligned}
\end{equation}
Using
$R_{k-1}^{(j)}=R_k^{(j)}+\Delta F_1^{(j)}(k)$, we obtain
\begin{equation} \nonumber
\begin{aligned}
\widetilde B_k^{(i)}
&=
\frac{1}{|\mathcal C_k^-(i)|}
\sum_{j\in \mathcal C_k^-(i)}
\Delta F_1^{(j)}(k) \\
&\quad+
\left[
\frac{1}{|\mathcal C_k^-(i)|}
\sum_{j\in \mathcal C_k^-(i)}
R_k^{(j)}
-
\frac{1}{|\mathcal C_k^+(i)|}
\sum_{j\in \mathcal C_k^+(i)}
R_k^{(j)}
\right].
\end{aligned}
\label{eq:app-cluster-shift}
\end{equation}

Here, the first term is the desired local increment baseline. 
The second term is introduced only because the two residual expectations are computed over different sets of rollouts.
This term is not part of the local effect of adding $e_k$, but the change in cluster membership between the two estimates, leading to cluster-reassignment bias. 
Using the fixed pre-element cluster $\mathcal C_k(i)$ for both residual terms eliminates this extra term and makes the residual difference a paired comparison with rollouts of similar quality.
We empirically verify the effectiveness of cluster reuse in Appendix~\ref{app:elerpo_results}.

\subsection{Empirical Validation of the Prefix-Clustering Approximation}
\label{app:theory:cluster-empirical}

We empirically validate that $\mathbb{E}[F_1{-}F_1(k)|e_{\leq k}, I] \approx \mathbb{E}[F_1{-}F_1(k)|F_1(k), I]$ by comparing the estimation based on this approximation against a re-rollout-based Monte-Carlo gold standard.
Specifically, we sample $500$ images from the held-out \dataset{} test set and evaluate at three layers $k \in \{n/4,\, n/2,\, 3n/4\}$, where $n$ is the number of ground-truth elements.
For each $(\text{image}, k)$ pair, we sample $80$ rollouts and retain those with at least $k$ elements.
For each retained rollout, we run $20$ independent re-rollouts from layer $k$ and obtain a gold-standard estimation $\tfrac{1}{20}\sum_{r=1}^{20} \big(F_1^{(r)} - F_1(k)\big)$.
This process yields $36{,}018$ rollouts and roughly $720$K scored re-rollouts.
We compare the gold standard with the estimate obtained by partitioning the rollouts into $2$ clusters using k-means on $F_1(k)$ and averaging $F_1{-}F_1(k)$ within each cluster.
Pooled across all $36{,}018$ rollouts, this estimate achieves a Pearson correlation of $0.811$ with the gold standard.
The $95\%$ confidence interval obtained by clustered bootstrap over the $500$ images is  $[0.787,\, 0.832]$.
This supports the approximation used in Eq.~(\ref{eq:method2}).

\subsection{Magnitude Analysis of the Element-Level Reward}
\label{app:theory:magnitude}

We prove that the magnitude of $\hat{r}_k^{(i)}$ in Eq.~(\ref{eq:method2}) is bounded by $\mathcal O(1/(k+n))$, where $n$ is the number of ground-truth elements.
Let $S_k^{(i)}=\sum_{i=1}^n s_i$ denote the total matched score between the $k$ lower-layer elements of rollout $i$ and the $n$ ground-truth elements. 
The corresponding $F_1$ score can be written as 

$$
F_1^{(i)}(k)=\frac{
2\left(S_k^{(i)}/k\right)\left(S_k^{(i)}/n\right)
}{
S_k^{(i)}/k+S_k^{(i)}/n
}
=\frac{2S_k^{(i)}}{k+n}.
$$

Let

$$
\delta_k^{(i)}=S_k^{(i)}-S_{k-1}^{(i)},
$$

we have

$$
0\le \delta_k^{(i)}\le 1.
$$

The lower bound follows because the previous matching remains feasible after adding a new prediction. 
The upper bound follows because the new prediction can participate in at most one matched pair, whose score is at most $1$.
Using the definition of $F_1$, the single-element increment is

$$
\Delta F_1^{(i)}(k)
=
\frac{2S_k^{(i)}}{k+n}
-
\frac{2S_{k-1}^{(i)}}{k+n-1}.
$$

Substituting $S_k^{(i)} = S_{k-1}^{(i)} + \delta_k^{(i)}$, we obtain

$$
\Delta F_1^{(i)}(k)
=
\frac{
2\big((k+n-1)\delta_k^{(i)}-S_{k-1}^{(i)}\big)
}{
(k+n)(k+n-1)
}.
$$

Since $0\le \delta_k^{(i)}\le 1$ and $0\le S_{k-1}^{(i)}\le n$, we have

$$
\left|\Delta F_1^{(i)}(k)\right|
\le
\frac{
2\big((k+n-1)+n\big)
}{
(k+n)(k+n-1)
}
\le
\frac{4}{k+n}.
$$

Thus,

$$
\left|\Delta F_1^{(i)}(k)\right|
=
\mathcal O\!\left(\frac{1}{k+n}\right).
$$

Finally, recall that

$$
\hat{r}_k^{(i)}
=
\Delta F_1^{(i)}(k)
-
\frac{1}{|\mathcal C_k(i)|}
\sum_{j\in \mathcal C_k(i)}
\Delta F_1^{(j)}(k).
$$

By the triangle inequality,

$$
\left|\hat r_k^{(i)}\right|
\le
\left|\Delta F_1^{(i)}(k)\right|
+
\frac{1}{|\mathcal C_k(i)|}
\sum_{j\in \mathcal C_k(i)}
\left|\Delta F_1^{(j)}(k)\right|
\le
\frac{8}{k+n}.
$$

Therefore,

$$
\left|\hat r_k^{(i)}\right|
=
\mathcal O\!\left(\frac{1}{k+n}\right).
$$

This layer-dependent scale motivates the layer-wise normalization in
Eq.~(\ref{eq:normalized-advantage}).

\section{\dataset{} Dataset Construction}
\label{app:dataset}

\begin{figure}[H]
    \centering
    \includegraphics[width=\linewidth]{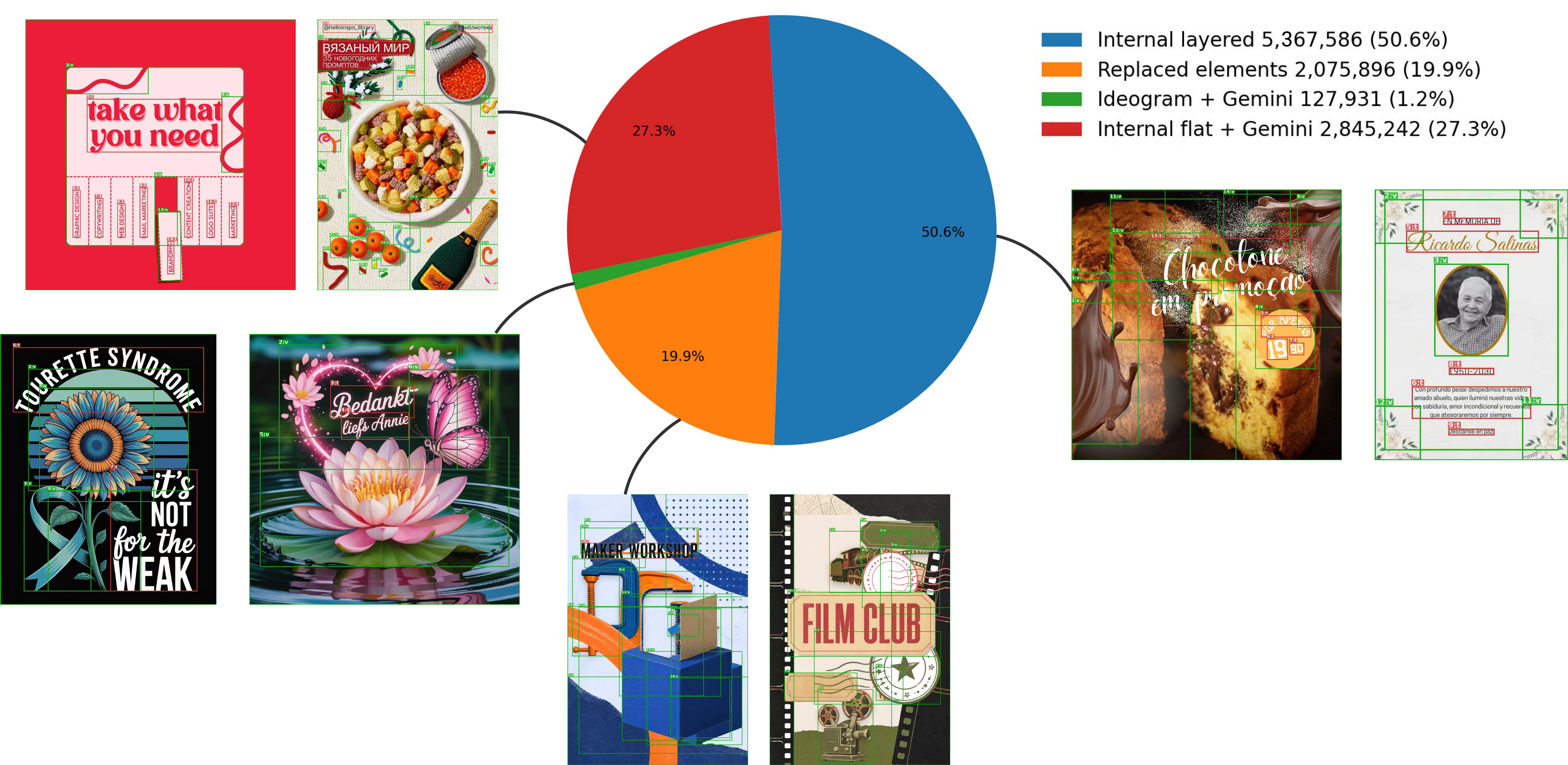}
    \caption{Distribution and examples of data in \dataset{}.}
    \label{fig:dataset-source-pie}
\end{figure}

As shown in Fig.~\ref{fig:dataset-source-pie}, our dataset is constructed from five sources: internal layered designs, designs created by replacing elements, designs generated using Ideogram, and internal flat designs.
Designs generated using Ideogram are obtained in accordance with its Terms of Service~\cite{ideogram_tos}, which specifies that users retain rights to generated outputs, including for commercial use.

For flat designs without original layer annotations, we use Gemini-3-Pro to annotate amodal bounding boxes and compositional orders.
The prompt used for annotation is provided below.

\begin{tcblisting}{
    colframe=cyan,
    colback=cyan!5,
    breakable,
    listing only,
    listing options={
        basicstyle=\ttfamily\small,
        breaklines=true,
        breakatwhitespace=true,
        columns=fullflexible,
        keepspaces=true
    }
}
Group this image into SEMANTIC layers.

Grouping rules:
- All text content -> group into logical text blocks (title, body, caption)
- Decorative elements -> group by visual similarity or region
- Each main subject -> one layer
- Do NOT include the background

IMPORTANT: Text and its background/underlay are SEPARATE layers!
- If text sits on a colored shape/box, that shape is a separate layer
- Text box = 2 layers: the background shape + the text itself

IMPORTANT: For ALL objects, return the FULL object box, not just the visible part.
- If a person, icon, illustration, photo cutout, decorative object, or other visual layer is partially occluded, the bbox should cover the object's full extent as best as you can infer it.
- Include hidden or covered parts in the bbox when the object is clearly partially blocked by another layer.

Think: "What logical groups would a designer work with?"

Output layers in back-to-front order (backmost first, frontmost last).

Return: {"layers": [{"type": "t" or "v", "bbox": [x1,y1,x2,y2]}]}
- "t" = text, "v" = visual (icon, image, shape, underlay, etc.)
- Coordinates normalized to [0,1].
\end{tcblisting}

Despite the availability of layer annotations, the model still faces granularity ambiguity: semantically independent elements may be merged into one (\eg, a cluster of decorative stars), while a single coherent element may be split into multiple parts (\eg, a logo decomposed into several decorative shapes).
We address this issue with a two-step filtering process.

First, we mine examples that emphasize small independent elements.
We train a preliminary detector on approximately $1$M automatically annotated internal designs.
We define small elements as elements whose normalized area is between $1\%$ and $5\%$ of the image area.
We retain designs unused in the training of the preliminary model that contain $5$--$10$ such small elements and for which the preliminary detector obtains a small-element recall between $0\%$ and $20\%$.
This step selects hard examples where small elements are likely to be missed or merged.
It produces approximately $38$K candidate designs.

Second, we remove designs that are likely to contain over-split annotations.
We detect pairs of boxes with non-contained overlap, where two boxes intersect but neither box fully contains the other.
We then group all boxes connected by such overlaps using union-find.
For each connected component, we compute the number of boxes and the fill ratio, defined as the sum of box areas divided by the area of the component's enclosing box.
We filter out designs containing an overlap component with at least $10$ boxes and a fill ratio greater than $2.0$.
These cases often correspond to a single coherent element split into many overlapping parts.
After filtering, we obtain $24{,}195$ high-quality designs.
We use $19{,}195$ designs for RL training and $5{,}000$ designs for testing.

\section{\model{} Training Details}
\label{app:impl}

We build \model{} upon Qwen3-VL-2B~\citep{bai2025qwen3}. 
Each element is represented by a bounding box, whose coordinates are normalized to $[0, 1000]$, and a category label indicating whether the element is text or visual. 
The model outputs a semicolon-delimited sequence where each element is encoded as $(x1, y1, x2, y2, c)$, with $(x1, y1, x2, y2)$ denoting the bounding box coordinates and $c \in \{t, v\}$ the category label, \eg, $102,35,530,210,v;45,680,980,750,t;$. 
We train the model with LoRA~\citep{hu2022lora} using rank $32$ and $\alpha=64$ across all linear layers. 
All training is conducted on NVIDIA H200 GPUs in bfloat16 precision, using AdamW~\citep{loshchilov2018decoupled} with $(\beta_1, \beta_2) = (0.9, 0.95)$ and weight decay $0.1$.

Training follows a two-stage pipeline. 
We first perform supervised fine-tuning (SFT) on 16 H200 GPUs for $650{,}000$ steps (${\sim}247$ hours) with a cosine learning rate schedule, a peak learning rate of $1\times10^{-4}$, a warmup ratio of $0.01$, and a per-device batch size of $4$. 
This stage uses all sources except the RL training split and the test split.

This is followed by RL using \alg{} on 8 H200 GPUs for $5{,}000$ steps (${\sim}20$ hours), with $16$ rollouts per image, a per-device batch size of $2$, a cosine learning rate schedule with a peak learning rate of $5\times10^{-7}$, and a warmup ratio of $0.01$. 
We use vLLM~\citep{kwon2023efficient} in colocated mode for efficient rollout generation, with a sampling temperature of $0.7$, top-$p$ of $0.9$, and top-$k$ of $20$. 
The KL regularization coefficient $\beta$ is set to $0.04$. 
Following Liu~\etal~\citep{liu2026gdpo}, we compute separate IoU-weighted $F_1$ rewards for text and visual elements, and use their average as the final reward.

\section{Detailed Experimental Setup for \model{} Evaluation} 
\label{app:eval}

\subsection{Dataset Statistics}
\label{app:eval:benchmarks}

We evaluate \model{} on the test split of \dataset{} and five external graphic design benchmarks.
To ensure consistency across benchmarks, we convert all annotations into text and visual elements.

\dataset{} contains $5{,}000$ held-out test designs from our high-quality subset.

Design-Multi-Layer-Bench~\cite{pu2025art} is a multi-layer graphic design benchmark introduced by ART~\cite{pu2025art}, containing $5{,}000$ designer-created works with original layered representations.

Crello~\cite{yamaguchi2021canvasvae} is a vector graphic document dataset collected from online design templates, and we use its official test split with $2{,}278$ designs.

PrismLayersPro~\cite{chen2025prismlayers} is a high-quality subset of PrismLayers~\cite{chen2025prismlayers}, containing $20{,}000$ multi-layer transparent images with alpha mattes.

LICA~\cite{hirsch2026lica} provides hierarchical component-level annotations for graphic designs, and we use the publicly released $1{,}148$ layouts and re-render images from the provided annotations.

ChartGalaxy~\cite{li2025chartgalaxy} is a large-scale infographic chart dataset with structured generation specifications, and we reconstruct layered representations of $100,000$ infographic charts using its generation pipeline and sample $2{,}000$ images for evaluation.
The infographic charts with layered representations are released on Huggingface.

We check each external benchmark independently against \dataset{} to ensure no duplicates.
For each benchmark, we first compare exact image hashes to identify identical samples.
No identical samples are identified in this step.
We then perform a CLIP-based near-duplicate check.
For each external benchmark, we encode both benchmark images and \dataset{} images using CLIP~\cite{radford2021learning}, rank all cross-dataset image pairs by cosine similarity, and manually inspect the top $1{,}000$ most similar pairs.
No duplicates are found.

\subsection{Baseline Implementations}
\label{app:eval:baselines}
We compare \model{} with a set of representative proprietary and open-source models.
For proprietary models, we use GPT-4o~\cite{gpt4o}, GPT-5~\cite{gpt5}, GPT-5.4~\cite{gpt5_4}, GPT-5.5~\cite{gpt5_5}, and Gemini-3.1-Pro~\cite{gemini3_1pro}.
For open-source models, we evaluate GLIP-L~\cite{li2022grounded}, Detic-B~\cite{zhou2022detecting}, Grounding DINO-B~\cite{liu2024grounding}, Florence2-L~\cite{xiao2024florence}, Qwen3-VL~\cite{bai2025qwen3}, InternVL3.5~\cite{wang2025internvl3}, and Rex-Omni~\cite{jiang2025rexomni}.

For proprietary models (GPT-4o, GPT-5, GPT-5.4, GPT-5.5, Gemini-3.1-Pro), we query each model's official API with the prompt described in Appendix~\ref{app:dataset}. 

For open-source vision-language models (Qwen3-VL-2B/8B, InternVL3.5-2B/8B), we use vLLM for accelerated batch inference.
These models receive the same structured prompt in Appendix~\ref{app:dataset}.

For traditional open-vocabulary detectors (GLIP-L, Detic-B, Grounding DINO-B, Florence2-L) and Rex-Omni, we use their official inference pipelines with the text prompt ["text element", "visual element"] as category queries. 
A confidence threshold of 0.3 is applied to GLIP-L predictions to filter low-quality detections.

All models are evaluated in a fully zero-shot setting without any task-specific fine-tuning or in-context examples.

\subsection{Human Evaluation Details}
\label{app:human}

\begin{figure}[H]
    \centering
    \includegraphics[width=0.85\linewidth]{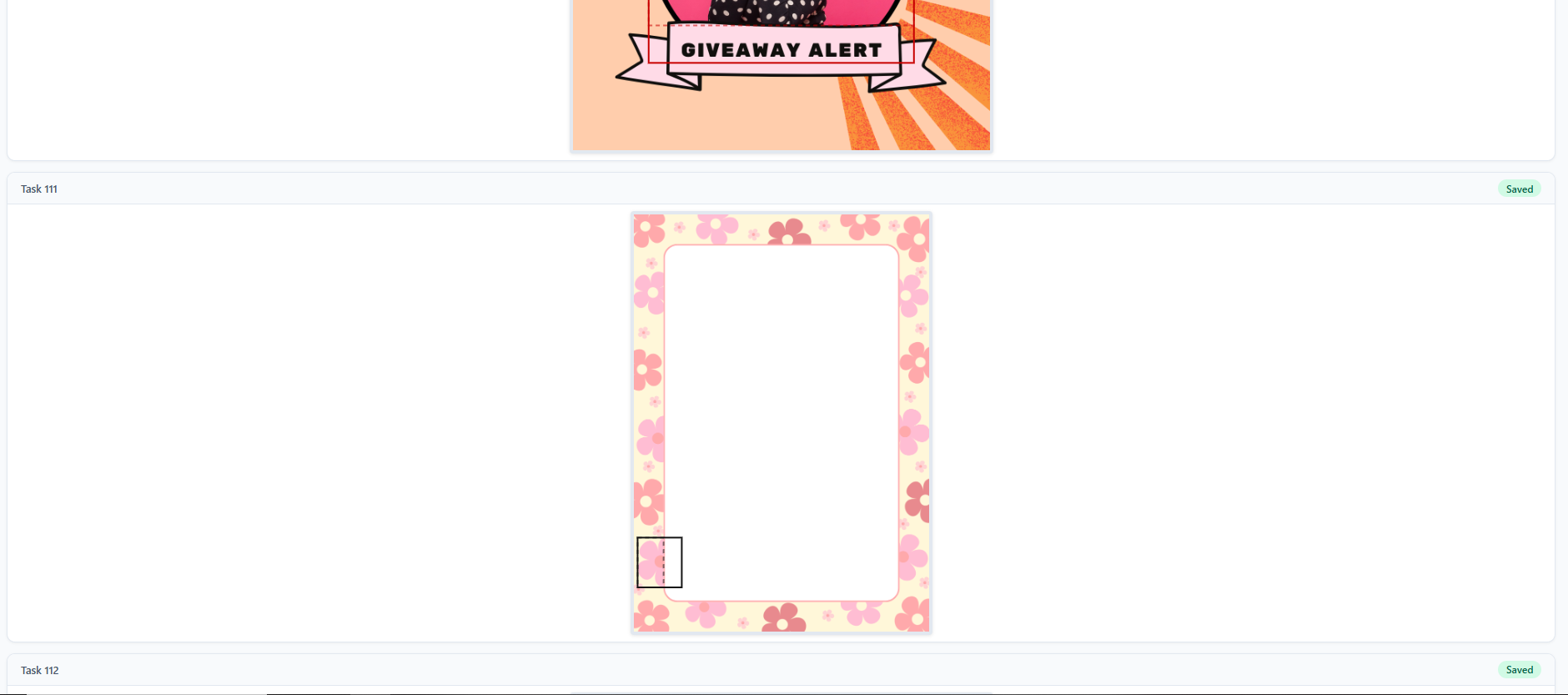}
    \caption{
    Annotation interface used in the human evaluation.
    }
    \label{fig:human_interface}
\end{figure}

We conduct human evaluation on amodal bounding-box prediction for occluded graphic elements.
Figure~\ref{fig:human_interface} shows the web-based annotation interface.
For each example, annotators are shown the visible region of a target occluded element and are asked to draw its amodal bounding box.
Before annotation, annotators are given a short instruction session with examples to clarify the task.
Model predictions are not shown to annotators.

Before participating, annotators read and signed an informed consent form.
The consent form explains the study purpose, the annotation procedure, and the estimated study duration.
It also states that participation is voluntary, that participants may stop the study at any time without providing a reason, and that they may request deletion of their data after the study.
All annotation data used in our analysis are anonymized and used only for research purposes.
Each annotator is compensated $\$15$ for completing the study, which takes approximately $20$ minutes.

\section{Additional Results for \model{} Evaluation}
\label{app:eval_results}

Tables~\ref{tab:sota}, \ref{tab:vc5k}, \ref{tab:crello}, \ref{tab:prismlayerspro}, \ref{tab:lica_dataset}, and \ref{tab:occlusion_test} compare \model{} with baselines.
We evaluate the models over $3$ runs and report their mean results, except for proprietary models, which are evaluated once due to the cost of API calls.
\model{} achieves the best performance across all six datasets, improving overall detection quality, amodal detection quality, and ordering accuracy.
It is also more efficient than proprietary baselines and comparable to open-source ones.

\begin{table}[H]
\centering
\caption{Evaluation results on \dataset{} test set.}
\label{tab:sota}
\tablestyle{3pt}{1.3}
\resizebox{0.9\linewidth}{!}{
\begin{tabular}{l|cc|c|c|c}
    \shline
    Method & $F_1$@[.5:.95]$\uparrow$ & IoU-weighted $F_1$$\uparrow$ & Occluded R@[.5:.95]$\uparrow$ & POA$\uparrow$ & Time (s)$\downarrow$ \\
    \hline
    Proprietary & & & & & \\
    \quad GPT-4o             & 4.16 & 7.56 & 5.83 & 1.88 & 2.40 \\
    \quad GPT-5              & 8.99 & 15.99 & 12.59 & 7.14 & 49.02 \\
    \quad GPT-5.4            & 22.41 & 33.37 & 23.31 & 13.94 & 3.16 \\
    \quad GPT-5.5            & 40.89 & 49.46 & 34.26 & 25.42 & 22.01 \\
    \quad Gemini-3.1-Pro     & 32.58 & 39.95 & 25.15 & 16.56 & 22.88 \\
    \hline
    Open-source & & & & & \\
    \quad GLIP-L             & 16.88$\pm$0.78 & 21.04$\pm$1.00 & 22.87$\pm$6.05 & - & 0.34 \\
    \quad Detic-B            & 4.77$\pm$0.00 & 5.16$\pm$0.00 & 2.47$\pm$0.00 & - & \textbf{0.19} \\
    \quad Grounding DINO-B   & 13.88$\pm$0.00 & 15.03$\pm$0.00 & 15.63$\pm$0.00 & - & 0.54 \\
    \quad Florence2-L        & 0.31$\pm$0.44 & 0.37$\pm$0.52 & 0.04$\pm$0.05 & - & 0.54 \\
    \quad Qwen3-VL-2B        & 11.75$\pm$0.01 & 15.71$\pm$0.01 & 6.26$\pm$0.00 & 1.03$\pm$0.00 & 0.86 \\
    \quad Qwen3-VL-8B        & 12.90$\pm$0.00 & 17.89$\pm$0.00 & 10.29$\pm$0.00 & 3.88$\pm$0.00 & 2.19 \\
    \quad InternVL3.5-2B     & 0.81$\pm$0.01 & 1.58$\pm$0.01 & 0.96$\pm$0.01 & 0.05$\pm$0.00 & 1.53 \\
    \quad InternVL3.5-8B     & 4.07$\pm$0.05 & 7.14$\pm$0.11 & 3.98$\pm$0.09 & 0.79$\pm$0.03 & 1.79 \\
    \quad Rex-Omni           & 31.24$\pm$0.03 & 38.12$\pm$0.04 & 15.41$\pm$0.00 & - & 0.51 \\
    \hline
    \rowcolor{blue!10} \model{} (Ours) & \textbf{72.57$\pm$0.04} & \textbf{79.65$\pm$0.03} & \textbf{65.35$\pm$0.13} & \textbf{50.18$\pm$0.53} & 0.65 \\
    \shline
\end{tabular}
}
\end{table}

\begin{table}[H]
\centering
\caption{Evaluation results on Design-Multi-Layer-Bench.}
\label{tab:vc5k}
\tablestyle{3pt}{1.3}
\resizebox{0.9\linewidth}{!}{
\begin{tabular}{l|cc|c|c}
    \shline
    Method & $F_1$@[.5:.95]$\uparrow$ & IoU-weighted $F_1$$\uparrow$ & Occluded R@[.5:.95]$\uparrow$ & POA$\uparrow$ \\
    \hline
    Proprietary & & & & \\
    \quad GPT-4o & 7.67 & 13.95 & 11.89 & 3.38 \\
    \quad GPT-5 & 13.58 & 24.21 & 21.11 & 8.74 \\
    \quad GPT-5.4 & 38.61 & 54.71 & 46.28 & 25.70 \\
    \quad GPT-5.5 & 61.05 & 70.24 & 61.35 & 34.17 \\
    \quad Gemini-3.1-Pro & 44.17 & 52.60 & 42.68 & 22.57 \\
    \hline
    Open-source & & & & \\
    \quad GLIP-L & 19.48$\pm$4.19 & 24.52$\pm$5.88 & 25.66$\pm$7.33 & - \\
    \quad Detic-B & 2.98$\pm$0.00 & 3.21$\pm$0.00 & 2.00$\pm$0.00 & - \\
    \quad Grounding DINO-B & 17.09$\pm$0.00 & 19.09$\pm$0.00 & 20.92$\pm$0.00 & - \\
    \quad Florence2-L & 0.01$\pm$0.00 & 0.02$\pm$0.00 & 0.02$\pm$0.00 & - \\
    \quad Qwen3-VL-2B & 29.60$\pm$0.00 & 37.21$\pm$0.00 & 19.36$\pm$0.00 & 7.79$\pm$0.00 \\
    \quad Qwen3-VL-8B & 38.06$\pm$0.00 & 49.21$\pm$0.00 & 32.80$\pm$0.00 & 16.01$\pm$0.00 \\
    \quad InternVL3.5-2B & 2.04$\pm$0.00 & 4.12$\pm$0.01 & 2.73$\pm$0.02 & 0.20$\pm$0.00 \\
    \quad InternVL3.5-8B & 9.95$\pm$0.04 & 17.46$\pm$0.07 & 11.86$\pm$0.10 & 3.44$\pm$0.03 \\
    \quad Rex-Omni & 17.79$\pm$0.00 & 19.19$\pm$0.00 & 13.25$\pm$0.00 & - \\
    \hline
    \rowcolor{blue!10} \model{} (Ours) & \textbf{79.52$\pm$0.09} & \textbf{84.42$\pm$0.14} & \textbf{71.65$\pm$0.23} & \textbf{46.97$\pm$0.22} \\
    \shline
\end{tabular}
}
\end{table}

\begin{table}[H]
\centering
\caption{Evaluation results on Crello.}
\label{tab:crello}
\tablestyle{3pt}{1.3}
\resizebox{0.9\linewidth}{!}{
\begin{tabular}{l|cc|c|c}
    \shline
    Method & $F_1$@[.5:.95]$\uparrow$ & IoU-weighted $F_1$$\uparrow$ & Occluded R@[.5:.95]$\uparrow$ & POA$\uparrow$ \\
    \hline
    Proprietary & & & & \\
    \quad GPT-4o & 5.48 & 9.64 & 10.64 & 1.95 \\
    \quad GPT-5 & 10.94 & 18.80 & 21.24 & 7.66 \\
    \quad GPT-5.4 & 23.15 & 33.43 & 36.50 & 15.22 \\
    \quad GPT-5.5 & 45.86 & 56.53 & 52.26 & 29.31 \\
    \quad Gemini-3.1-Pro & 34.32 & 42.20 & 36.91 & 18.77 \\
    \hline
    Open-source & & & & \\
    \quad GLIP-L & 16.60$\pm$4.51 & 22.15$\pm$6.30 & 47.45$\pm$6.11 & - \\
    \quad Detic-B & 5.97$\pm$0.00 & 6.57$\pm$0.00 & 8.02$\pm$0.00 & - \\
    \quad Grounding DINO-B & 12.29$\pm$0.00 & 14.19$\pm$0.00 & 21.33$\pm$0.00 & - \\
    \quad Florence2-L & 0.05$\pm$0.00 & 0.07$\pm$0.00 & 0.08$\pm$0.00 & - \\
    \quad Qwen3-VL-2B & 12.32$\pm$0.00 & 17.17$\pm$0.00 & 9.82$\pm$0.00 & 2.18$\pm$0.00 \\
    \quad Qwen3-VL-8B & 20.04$\pm$0.00 & 27.33$\pm$0.00 & 20.05$\pm$0.00 & 8.60$\pm$0.00 \\
    \quad InternVL3.5-2B & 1.64$\pm$0.01 & 3.05$\pm$0.02 & 2.82$\pm$0.04 & 0.15$\pm$0.00 \\
    \quad InternVL3.5-8B & 5.45$\pm$0.54 & 9.45$\pm$1.05 & 8.97$\pm$0.38 & 0.86$\pm$0.25 \\
    \quad Rex-Omni & 26.01$\pm$0.02 & 33.93$\pm$0.03 & 22.34$\pm$0.02 & - \\
    \hline
    \rowcolor{blue!10} \model{} (Ours) & \textbf{57.75$\pm$0.14} & \textbf{68.31$\pm$0.23} & \textbf{58.03$\pm$0.20} & \textbf{34.89$\pm$0.39} \\
    \shline
\end{tabular}
}
\end{table}

\begin{table}[H]
\centering
\caption{Evaluation results on PrismLayersPro.}
\label{tab:prismlayerspro}
\tablestyle{3pt}{1.3}
\resizebox{0.9\linewidth}{!}{
\begin{tabular}{l|cc|c|c}
    \shline
    Method & $F_1$@[.5:.95]$\uparrow$ & IoU-weighted $F_1$$\uparrow$ & Occluded R@[.5:.95]$\uparrow$ & POA$\uparrow$ \\
    \hline
    Proprietary & & & & \\
    \quad GPT-4o & 5.44 & 11.27 & 8.54 & 2.09 \\
    \quad GPT-5 & 11.71 & 23.44 & 18.55 & 7.52 \\
    \quad GPT-5.4 & 39.34 & 61.12 & 42.78 & 29.94 \\
    \quad GPT-5.5 & 61.41 & 73.66 & 60.53 & 37.29 \\
    \quad Gemini-3.1-Pro & 40.75 & 48.54 & 37.71 & 22.32 \\
    \hline
    Open-source & & & & \\
    \quad GLIP-L & 24.30$\pm$6.80 & 29.71$\pm$8.83 & 26.45$\pm$7.97 & - \\
    \quad Detic-B & 9.96$\pm$0.00 & 10.53$\pm$0.00 & 6.62$\pm$0.00 & - \\
    \quad Grounding DINO-B & 28.28$\pm$0.00 & 30.31$\pm$0.00 & 32.37$\pm$0.00 & - \\
    \quad Florence2-L & 0.19$\pm$0.14 & 0.25$\pm$0.17 & 0.27$\pm$0.19 & - \\
    \quad Qwen3-VL-2B & 37.13$\pm$0.00 & 45.19$\pm$0.00 & 24.11$\pm$0.00 & 11.55$\pm$0.00 \\
    \quad Qwen3-VL-8B & 49.35$\pm$0.02 & 61.04$\pm$0.00 & 39.22$\pm$0.10 & 26.39$\pm$0.06 \\
    \quad InternVL3.5-2B & 1.19$\pm$0.00 & 2.60$\pm$0.00 & 1.70$\pm$0.00 & 0.09$\pm$0.00 \\
    \quad InternVL3.5-8B & 6.77$\pm$0.02 & 13.70$\pm$0.03 & 8.18$\pm$0.03 & 2.11$\pm$0.00 \\
    \quad Rex-Omni & 35.14$\pm$0.00 & 38.29$\pm$0.00 & 25.42$\pm$0.00 & - \\
    \hline
    \rowcolor{blue!10} \model{} (Ours) & \textbf{73.21$\pm$0.04} & \textbf{78.44$\pm$0.04} & \textbf{72.67$\pm$0.04} & \textbf{43.87$\pm$0.04} \\
    \shline
\end{tabular}
}
\end{table}

\begin{table}[H]
\centering
\caption{Evaluation results on LICA.}
\label{tab:lica_dataset}
\tablestyle{3pt}{1.3}
\resizebox{0.9\linewidth}{!}{
\begin{tabular}{l|cc|c|c}
    \shline
    Method & $F_1$@[.5:.95]$\uparrow$ & IoU-weighted $F_1$$\uparrow$ & Occluded R@[.5:.95]$\uparrow$ & POA$\uparrow$ \\
    \hline
    Proprietary & & & & \\
    \quad GPT-4o & 4.03 & 7.53 & 6.91 & 2.30 \\
    \quad GPT-5 & 8.83 & 15.99 & 14.88 & 7.94 \\
    \quad GPT-5.4 & 26.10 & 37.07 & 33.23 & 17.96 \\
    \quad GPT-5.5 & 45.92 & 54.95 & 45.10 & 29.53 \\
    \quad Gemini-3.1-Pro & 39.42 & 47.90 & 35.53 & 22.17 \\
    \hline
    Open-source & & & & \\
    \quad GLIP-L & 18.47$\pm$2.88 & 23.54$\pm$3.57 & 44.71$\pm$7.44 & - \\
    \quad Detic-B & 4.63$\pm$0.00 & 4.90$\pm$0.00 & 3.63$\pm$0.00 & - \\
    \quad Grounding DINO-B & 15.20$\pm$0.00 & 16.48$\pm$0.00 & 19.71$\pm$0.00 & - \\
    \quad Florence2-L & 0.12$\pm$0.00 & 0.15$\pm$0.00 & 0.18$\pm$0.00 & - \\
    \quad Qwen3-VL-2B & 14.79$\pm$0.00 & 19.28$\pm$0.00 & 10.87$\pm$0.00 & 1.60$\pm$0.00 \\
    \quad Qwen3-VL-8B & 19.22$\pm$0.01 & 26.27$\pm$0.02 & 17.75$\pm$0.00 & 6.70$\pm$0.00 \\
    \quad InternVL3.5-2B & 0.88$\pm$0.03 & 1.82$\pm$0.04 & 1.22$\pm$0.04 & 0.11$\pm$0.01 \\
    \quad InternVL3.5-8B & 6.11$\pm$0.08 & 10.40$\pm$0.08 & 6.98$\pm$0.06 & 1.93$\pm$0.01 \\
    \quad Rex-Omni & 28.51$\pm$0.15 & 35.70$\pm$0.22 & 16.37$\pm$0.01 & - \\
    \hline
    \rowcolor{blue!10} \model{} (Ours) & \textbf{71.02$\pm$2.76} & \textbf{78.89$\pm$3.35} & \textbf{62.67$\pm$0.66} & \textbf{46.96$\pm$1.19} \\
    \shline
\end{tabular}
}
\end{table}

\begin{table}[H]
\centering
\caption{Evaluation results on ChartGalaxy.}
\label{tab:occlusion_test}
\tablestyle{3pt}{1.3}
\resizebox{0.9\linewidth}{!}{
\begin{tabular}{l|cc|c|c}
    \shline
    Method & $F_1$@[.5:.95]$\uparrow$ & IoU-weighted $F_1$$\uparrow$ & Occluded R@[.5:.95]$\uparrow$ & POA$\uparrow$ \\
    \hline
    Proprietary & & & & \\
    \quad GPT-4o & 0.39 & 0.95 & 0.29 & 0.04 \\
    \quad GPT-5 & 1.30 & 2.84 & 1.46 & 0.42 \\
    \quad GPT-5.4 & 6.08 & 9.47 & 3.99 & 1.70 \\
    \quad GPT-5.5 & 6.85 & 9.42 & 2.98 & 1.12 \\
    \quad Gemini-3.1-Pro & 12.29 & 17.95 & 9.27 & 6.92 \\
    \hline
    Open-source & & & & \\
    \quad GLIP-L & 12.64$\pm$4.45 & 17.71$\pm$6.52 & 9.82$\pm$4.40 & - \\
    \quad Detic-B & 5.73$\pm$0.00 & 6.63$\pm$0.00 & 2.01$\pm$0.00 & - \\
    \quad Grounding DINO-B & 3.32$\pm$0.00 & 3.84$\pm$0.00 & 2.30$\pm$0.00 & - \\
    \quad Florence2-L & 0.00$\pm$0.00 & 0.00$\pm$0.00 & 0.00$\pm$0.00 & - \\
    \quad Qwen3-VL-2B & 0.08$\pm$0.00 & 0.18$\pm$0.00 & 0.03$\pm$0.00 & 0.01$\pm$0.00 \\
    \quad Qwen3-VL-8B & 0.99$\pm$0.00 & 1.93$\pm$0.00 & 0.33$\pm$0.00 & 0.08$\pm$0.00 \\
    \quad InternVL3.5-2B & 0.08$\pm$0.00 & 0.19$\pm$0.00 & 0.03$\pm$0.00 & 0.00$\pm$0.00 \\
    \quad InternVL3.5-8B & 0.40$\pm$0.02 & 0.86$\pm$0.02 & 0.22$\pm$0.00 & 0.05$\pm$0.01 \\
    \quad Rex-Omni & 6.31$\pm$0.00 & 8.24$\pm$0.00 & 3.16$\pm$0.00 & - \\
    \hline
    \rowcolor{blue!10} \model{} (Ours) & \textbf{32.97$\pm$1.48} & \textbf{57.71$\pm$3.62} & \textbf{35.76$\pm$4.88} & \textbf{27.71$\pm$6.58} \\
    \shline
\end{tabular}
}
\end{table}

\begin{figure}[H]
    \centering
    \includegraphics[width=1\linewidth]{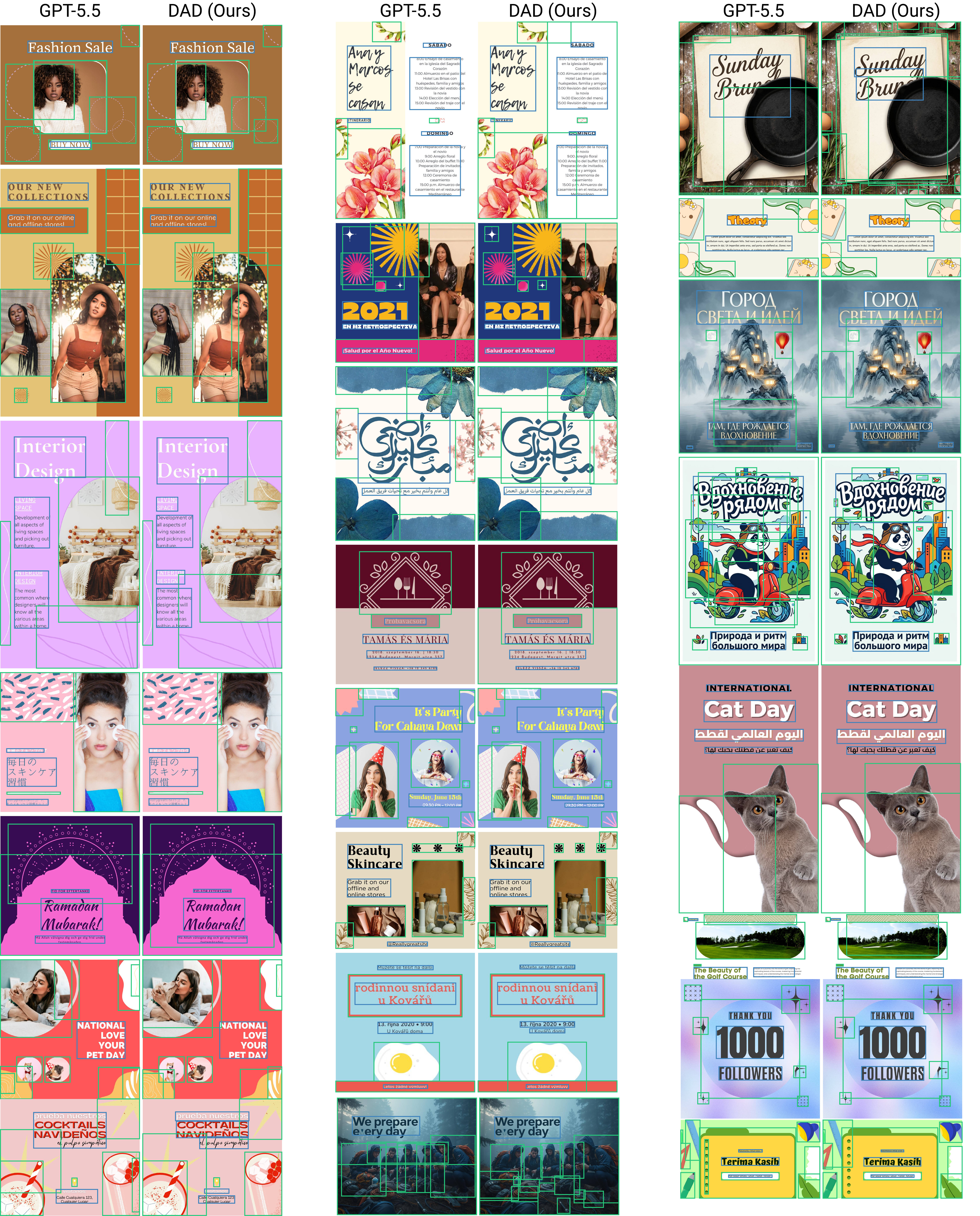}
    \caption{
      Additional qualitative comparison between \model{} and GPT-5.5.
    }
    \label{fig:app_qual}
\end{figure}

As shown in Fig.~\ref{fig:app_qual}, \model{} performs robustly across diverse designs, detecting elements missed by GPT-5.5 and recovering their occluded regions.

\section{Detailed Experimental Setup for \alg{} Evaluation} 
\label{app:elerpo}

\subsection{Hyperparameters of RL Baselines}
\label{app:elerpo:hparams}

We compare \alg{} with GRPO and its extensions that incorporate finer-grained supervision, including representative policy-signal-based methods (GRPO-$\lambda$~\cite{parthasarathi2025grpo}, GTPO~\cite{tan2025gtpo}, and PRIME~\cite{cui2025process}) and re-rollout-based methods (VinePPO~\cite{kazemnejad2025vineppo} and SPO~\cite{guo2026segment}).
Method-specific hyperparameters follow the original papers: \begin{itemize} 
    \item \textbf{GRPO-$\lambda$}: discount factor $\lambda{=}0.99$, return scaling $\gamma{=}1.0$, advantage clamp threshold $-0.1$, with both forward and backward $\lambda$-return variants applied. 
    \item \textbf{GTPO}: entropy-weighted advantage with scaling coefficients $\alpha_1{=}1.0$, $\alpha_2{=}0.1$, and entropy clipped to $[0.2, 0.28]$. 
    \item \textbf{PRIME}: process reward model (PRM) learning rate $1{\times}10^{-6}$, PRM update interval $1$ step, $K{=}4$ rollouts for PRM training, reward mixing coefficient $\beta_{\text{prime}}{=}0.05$, and $\gamma{=}1.0$. 
    \item \textbf{VinePPO}: $K{=}9$ re-rollouts per cutpoint, return scaling $\gamma{=}1.0$.
    \item \textbf{SPO}: chain mode with segment-level partitioning, $K{=}9$ Monte Carlo rollouts per segment, cutpoint threshold $0.9$, and segment-level KL coefficient $\beta_{\text{spo}}{=}0.0001$. 
\end{itemize}

\subsection{Element Order Experimental Setup}
\label{app:elerpo:order}
We compare three orders for predicting elements: random, left-edge order that sorts elements by their leftmost x-coordinate, and compositional order.
Since random and left-edge orderings destroy the implicit compositional order conveyed by the sequence position, we augment the output format for these two variants with an explicit zero-indexed order field, resulting in the representation $(x1, y1, x2, y2, c, index)$, \eg, $102,35,530,210,v,3;$. 
The compositional order variant retains the original format without an explicit index, as the order is naturally encoded by the sequence position.

For SFT, we train all three variants from the base Qwen3-VL-2B checkpoint for $31{,}250$ steps on 8 H200 GPUs with a learning rate of $1\times10^{-4}$ and an effective batch size of $32$, totaling approximately $1$M training samples. 
The RL stage uses the same configuration as specified in Appendix~\ref{app:impl}.

\subsection{Dataset Statistics}
\label{app:eval_ele:benchmarks}
We evaluate the generalizability of \alg{} across models and benchmarks.
The benchmarks span graphic layout benchmarks with underlying layer structure, those without explicit layer structure, and natural image benchmarks.
To ensure consistency across benchmarks, we convert all annotations into text and visual elements.

ChartGalaxy~\cite{li2025chartgalaxy} is a large-scale infographic chart dataset with structured chart generation specifications.
As described in Appendix~\ref{app:eval:benchmarks}, we reconstruct layered graphic designs from these specifications and sample $5{,}000$ images for training and $2{,}000$ images for evaluation.

Crello~\cite{yamaguchi2021canvasvae} is a vector graphic document dataset as described in Appendix~\ref{app:eval:benchmarks}.
We use its official train/test split.

Rico~\cite{deka2017rico} contains over $66$K user interfaces collected from Android applications.
Following common practice~\cite{manandhar2020learning,manandhar2021magic}, we split the dataset into $53$K layouts for training and $13$K for testing.

PosterLayout~\cite{hsu2023posterlayout} is a content-aware poster layout benchmark with annotated layout elements such as text, logos, and decorative shapes.
Following Zhu~\etal~\cite{zhu2026infodet}, we randomly split the dataset into $8,974$ for training and $1,000$ for testing

DocLayNet~\cite{pfitzmann2022doclaynet} is a document-layout analysis dataset with human-annotated bounding boxes for document components such as text, tables, figures, titles, and lists.
We use the official split with $69{,}375$ training pages and evaluate on the $6{,}489$ validation pages.

COCO~\cite{lin2014microsoft} is a natural-image object detection benchmark.
We use the train2017 split with $118{,}287$ images for training and the val2017 split with $5{,}000$ images for evaluation.

LVIS~\cite{gupta2019lvis} is a large-vocabulary detection benchmark with long-tailed object categories.
We use the official training split and evaluate on the validation split.

BDD100K~\cite{yu2020bdd100k} is a driving-scene benchmark with diverse weather and scene conditions.
We use the object detection split and report results on the validation set.

Objects365~\cite{shao2019objects365} is a large-scale natural-image object detection benchmark.
We use the official training split and evaluate on the validation split.

\section{Additional Results for \alg{} Evaluation}
\label{app:elerpo_results}

\begin{table}[H]
\centering
\caption{Additional ablation studies on key components of \alg{}.}
\label{tab:ablation_appendix}
\setlength{\tabcolsep}{4pt}
\renewcommand{\arraystretch}{1.2}
\resizebox{\linewidth}{!}{%
\begin{tabular}{@{}c@{\hspace{20pt}}c@{\hspace{20pt}}c@{}}
\begin{tabular}[t]{@{}c@{}}
(a) Mean Subtraction \\[3pt]
\begin{tabular}{cc}
    \shline
    Global & \cellcolor{blue!10}Layer \\
    \hline
    72.4 & \cellcolor{blue!10}\textbf{79.7} \\
    \shline
\end{tabular}
\end{tabular}
&
\begin{tabular}[t]{@{}c@{}}
(b) Std Division \\[3pt]
\begin{tabular}{ccc}
    \shline
    None & Global & \cellcolor{blue!10}Layer \\
    \hline
    74.0 & 78.6 & \cellcolor{blue!10}\textbf{79.7} \\
    \shline
\end{tabular}
\end{tabular}
&
\begin{tabular}[t]{@{}c@{}}
(c) Prefix Clustering \\[3pt]
\begin{tabular}{cccc}
    \shline
    Different cluster & None & k-means &  \cellcolor{blue!10}Silhouette \\
    \hline
    78.7 & 79.2 & 79.3 & \cellcolor{blue!10}\textbf{79.7} \\
    \shline
\end{tabular}
\end{tabular}
\end{tabular}
}
\end{table}

We present additional ablation studies on key components of \alg{}.

\paragraph{Mean subtraction} 
Table~\ref{tab:ablation_appendix}(a) replaces the per-step cluster-wise mean in Eq.~(\ref{eq:method2}) with a single global mean computed over all $\Delta F_1$ values across all element positions. 
This leads to a performance drop, confirming that conditioning on the element position $k$ is necessary.

\paragraph{Standard deviation division} Table~\ref{tab:ablation_appendix}(b) considers two alternatives to the per-step RMS normalization in Eq.~(\ref{eq:normalized-advantage}): 1)~removing it entirely, leaving the raw mean-subtracted rewards unnormalized; 
and 2)~replacing the per-step RMS with a global RMS pooled across all element positions.
Both variants degrade performance, confirming that per-step normalization is necessary to account for the different reward variance scales at different positions.

\paragraph{Prefix clustering} 
Table~\ref{tab:ablation_appendix}(c) compares four strategies for forming the cluster $\mathcal{C}_k(i)$ in Eq.~(\ref{eq:method2}): 
1)~separate clustering, which independently clusters on $F_1(k{-}1)$ and $F_1(k)$ to form different clusters $\mathcal{C}_k^-(i)$ and $\mathcal{C}_k^+(i)$ for the two residual terms;
2)~no clustering, where all rollouts share a single global mean; 
3)~fixed k-means with k=4, which always partitions rollouts into exactly 4 groups; and 
4)~our silhouette-based method. 
Our method performs best. 
The separate-clustering variant underperforms because using different clusters for the two residual terms introduces a cluster-reassignment bias.
Notably, even the no-clustering variant is already competitive, suggesting that the compositional order makes a single global mean a reasonable estimate, while adaptive clustering on prefix quality further refines this estimate.
%%%%%%%%%%%%%%%%%%%%%%%%%%%%%%%%%%%%%%%%%%%%%%%%%%%%%%%%%%%%

\newpage
\section*{NeurIPS Paper Checklist}

\begin{enumerate}

\item {\bf Claims}
    \item[] Question: Do the main claims made in the abstract and introduction accurately reflect the paper's contributions and scope?
    \item[] Answer: \answerYes{} % Replace by \answerYes{}, \answerNo{}, or \answerNA{}.
    \item[] Justification: Main claims in the abstract and introduction are backed by the task formulation, optimization method, and dataset construction in Sec.~\ref{sec:method}, and the experiments in Sec.~\ref{sec:exp}.

\item {\bf Limitations}
    \item[] Question: Does the paper discuss the limitations of the work performed by the authors?
    \item[] Answer: \answerYes{} % Replace by \answerYes{}, \answerNo{}, or \answerNA{}.
    \item[] Justification: We discuss the limitations as future work in Sec.~\ref{sec:conclusion}.

\item {\bf Theory assumptions and proofs}
    \item[] Question: For each theoretical result, does the paper provide the full set of assumptions and a complete (and correct) proof?
    \item[] Answer: \answerYes{} % Replace by \answerYes{}, \answerNo{}, or \answerNA{}.
    \item[] Justification: We provide the assumptions in Sec.~\ref{subsubsec:martingale} and the complete proof in Appendix~\ref{app:theory}.

    \item {\bf Experimental result reproducibility}
    \item[] Question: Does the paper fully disclose all the information needed to reproduce the main experimental results of the paper to the extent that it affects the main claims and/or conclusions of the paper (regardless of whether the code and data are provided or not)?
    \item[] Answer: \answerYes{} % Replace by \answerYes{}, \answerNo{}, or \answerNA{}.
    \item[] Justification: We describe how we build our model in Sec.~\ref{sec:method}, and provide experimental details in Appendices~\ref{app:impl}, \ref{app:eval} and \ref{app:elerpo}.

\item {\bf Open access to data and code}
    \item[] Question: Does the paper provide open access to the data and code, with sufficient instructions to faithfully reproduce the main experimental results, as described in supplemental material?
    \item[] Answer: \answerYes{} % Replace by \answerYes{}, \answerNo{}, or \answerNA{}.
    \item[] Justification:  We have released our code and $100,000$ designs in our dataset, and provided the links in the abstract.

\item {\bf Experimental setting/details}
    \item[] Question: Does the paper specify all the training and test details (e.g., data splits, hyperparameters, how they were chosen, type of optimizer) necessary to understand the results?
    \item[] Answer: \answerYes{} % Replace by \answerYes{}, \answerNo{}, or \answerNA{}.
    \item[] Justification: We have provided details for the experimental setup in Appendices~\ref{app:impl}, \ref{app:eval}, and \ref{app:elerpo}.

\item {\bf Experiment statistical significance}
    \item[] Question: Does the paper report error bars suitably and correctly defined or other appropriate information about the statistical significance of the experiments?
    \item[] Answer: \answerYes{} % Replace by \answerYes{}, \answerNo{}, or \answerNA{}.
    \item[] Justification: We report the variance for the evaluation of object detection models in Appendix~\ref{app:eval_results}.

\item {\bf Experiments compute resources}
    \item[] Question: For each experiment, does the paper provide sufficient information on the computer resources (type of compute workers, memory, time of execution) needed to reproduce the experiments?
    \item[] Answer: \answerYes{} % Replace by \answerYes{}, \answerNo{}, or \answerNA{}.
    \item[] Justification: We report the compute resources to conduct our experiments in Appendix~\ref{app:impl}.
    
\item {\bf Code of ethics}
    \item[] Question: Does the research conducted in the paper conform, in every respect, with the NeurIPS Code of Ethics \url{https://neurips.cc/public/EthicsGuidelines}?
    \item[] Answer: \answerYes{} % Replace by \answerYes{}, \answerNo{}, or \answerNA{}.
    \item[] Justification: We have read the NeurIPS Code of Ethics and made sure that the paper conforms to it.

\item {\bf Broader impacts}
    \item[] Question: Does the paper discuss both potential positive societal impacts and negative societal impacts of the work performed?
    \item[] Answer: \answerNA{} % Replace by \answerYes{}, \answerNo{}, or \answerNA{}.
    \item[] Justification: There are no obvious paths that this work leads to potential negative societal impacts.
    
\item {\bf Safeguards}
    \item[] Question: Does the paper describe safeguards that have been put in place for responsible release of data or models that have a high risk for misuse (e.g., pre-trained language models, image generators, or scraped datasets)?
    \item[] Answer: \answerNA{} % Replace by \answerYes{}, \answerNo{}, or \answerNA{}.
    \item[] Justification:  We do not foresee any high risk for misuse of this work.

\item {\bf Licenses for existing assets}
    \item[] Question: Are the creators or original owners of assets (e.g., code, data, models), used in the paper, properly credited and are the license and terms of use explicitly mentioned and properly respected?
    \item[] Answer: \answerYes{} % Replace by \answerYes{}, \answerNo{}, or \answerNA{}.
    \item[] Justification: We credit the original owners of assets and mention the terms of use in Appendix~\ref{app:dataset}.

\item {\bf New assets}
    \item[] Question: Are new assets introduced in the paper well documented and is the documentation provided alongside the assets?
    \item[] Answer: \answerYes{} % Replace by \answerYes{}, \answerNo{}, or \answerNA{}.
    \item[] Justification: We host our dataset on Huggingface and provide detailed documentation. The Huggingface link to the dataset is provided in the abstract.

\item {\bf Crowdsourcing and research with human subjects}
    \item[] Question: For crowdsourcing experiments and research with human subjects, does the paper include the full text of instructions given to participants and screenshots, if applicable, as well as details about compensation (if any)? 
    \item[] Answer: \answerYes{} % Replace by \answerYes{}, \answerNo{}, or \answerNA{}.
    \item[] Justification: We provide the screenshot of the web-based prototype and compensation details in Appendix~\ref{app:human}.

\item {\bf Institutional review board (IRB) approvals or equivalent for research with human subjects}
    \item[] Question: Does the paper describe potential risks incurred by study participants, whether such risks were disclosed to the subjects, and whether Institutional Review Board (IRB) approvals (or an equivalent approval/review based on the requirements of your country or institution) were obtained?
    \item[] Answer: \answerYes{} % Replace by \answerYes{}, \answerNo{}, or \answerNA{}.
    \item[] Justification: The paper describes the human evaluation protocol in Appendix~\ref{app:human}, including how consent was received from participants.
    The study involves only a simple, low-risk annotation task lasting approximately 20 minutes and has received IRB approval.

\item {\bf Declaration of LLM usage}
    \item[] Question: Does the paper describe the usage of LLMs if it is an important, original, or non-standard component of the core methods in this research? Note that if the LLM is used only for writing, editing, or formatting purposes and does \emph{not} impact the core methodology, scientific rigor, or originality of the research, declaration is not required.
    %this research? 
    \item[] Answer: \answerYes{} % Replace by \answerYes{}, \answerNo{}, or \answerNA{}.
    \item[] Justification: We describe how we build our detection model on top of a VLM, Qwen3-VL-2B, in Sec.~\ref{sec:method}.

\end{enumerate}

\end{document}